%% file: emnlp2022.tex
\pdfoutput=1
\documentclass[11pt]{article}

\usepackage[]{EMNLP2022}

\usepackage{times}
\usepackage{latexsym}

\usepackage[T1]{fontenc}
\usepackage[utf8]{inputenc}

\usepackage{microtype}

\usepackage{inconsolata}

\usepackage{microtype}
\usepackage{graphicx}
\usepackage{math_cmd}
\usepackage{xcolor}
\usepackage{algpseudocode}
\usepackage{algorithm}
\usepackage{breakcites}
\usepackage{pifont}
\usepackage{diagbox}
\usepackage{multirow}
\usepackage{booktabs}
\usepackage{enumitem}
\usepackage{bbm}
\usepackage{amsmath}
\usepackage{mathtools}

\usepackage{footnote}
\makesavenoteenv{tabular}

\title{Open World Classification with Adaptive Negative Samples}

\author{Ke Bai$^1$, Guoyin Wang$^2$, Jiwei Li$^3$, Sunghyun Park$^2$, \\
\textbf{Sungjin Lee$^2$, Puyang Xu$^2$, Ricardo Henao$^{1,4}$, Lawrence Carin$^4$} \\
  $^1$Duke University $^2$Amazon $^3$Zhejiang University $^4$KAUST\\
  \small\texttt{\{ke.bai, ricardo.henao\}@duke.edu, \{guoyiwan,  sunghyu, sungjinl, puyax\}@amazon.com,} \\
  \small\texttt{jiwei\_li@zju.edu.cn, larry.carin@kaust.edu.sa} \\
  }

\begin{document}
\maketitle
\begin{abstract}
Open world classification is a task in natural language processing with key practical relevance and impact.
% , \eg, out-of-scope intent detection.
Since the open or {\em unknown} category data only manifests in the inference phase, finding a model with a suitable decision boundary accommodating for the identification of known classes and discrimination of the open category is challenging.
The performance of existing models is limited by the lack of effective open category data during the training stage or the lack of a good mechanism to learn appropriate decision boundaries.
We propose an approach based on \underline{a}daptive \underline{n}egative \underline{s}amples (ANS) designed to generate effective synthetic open category samples in the training stage and without requiring any prior knowledge or external datasets.
Empirically, we find a significant advantage in using auxiliary one-versus-rest binary classifiers, which 
effectively utilize the generated negative samples and avoid the complex threshold-seeking stage in previous works.
% \textcolor{red}{We also found the great advantage of using one-versus-rest binary classifier, which can fully utilized our generated samples and avoid complex threshold seeking as the most previous works did.}
% We also propose a new open set recognition  framework to better utilize the synthetic data and learn robust classifiers without complex threshold seeking.
% We also propose a new decision boundary learning framework to utilize the synthetic data and learn robust classifiers without complex threshold seeking. 
% \bc{too redundant? Can consider delete the following sentence in green}
% \textcolor{green}{
% Specially, we design a robust local negative sample construction constraint and further utilize a gradient-ascend method to generate the samples. 
% We then couple a multi-class classifier and auxiliary binary one-vs-rest classifiers learned on both training and synthetic data to obtain a tight decision boundary.}
Extensive experiments on three benchmark datasets show that ANS achieves significant improvements over state-of-the-art methods.
\end{abstract}

\input{introduction}
\input{related}
\input{methods}
\input{experiments}
\input{conclusion}
\input{limitations}
\input{acknowledgement}
\bibliography{anthology,emnlp}
\bibliographystyle{acl_natbib}

\appendix
% \pagebreak
\clearpage
\input{supplementary}

% \section{Example Appendix}
% \label{sec:appendix}

% This is a section in the appendix.

\end{document}

%% file: introduction.tex
\section{Introduction}
% Intent classification is a supervised problem, identifying the users' intention to one of categories appeared in the training set. 
% Supervised classification assumes that categories of the testing sample have been revealed in the training phase. Every sample has to be categorized into one category (Fig.~\ref{fig:motivation}(a)). This may not always be satisfied in piratical applications, like dialogue intention classification, where new intents from the customer may emerge. The discriminator should be capable of discriminating whether this sample belongs to the known categories or an unseen category. We could regard this problem as a $(C+1)$ classification problem, where the ``$1$'' represents the unknown category. This problem is also known as multi-category open set
Standard supervised classification assumes that all categories expected in the testing phase have been fully observed while training, {\em i.e.}, every sample is assigned to one {\em known} category as illustrated in Figure~\ref{fig:motivation}(a).
This may not always be satisfied in practical applications, such as dialogue intention classification, where new intents are expected to emerge.
Consequently, it is desirable to have a classifier capable of discriminating whether a given sample belongs to a known or an unknown category, {\em e.g.}, the red samples in Figure~\ref{fig:motivation}(a).
This problem can be understood as a $(C+1)$ classification problem, where $C$ is the number of known categories and the additional category is reserved for {\em open} (unknown) samples.
This scenario is also known as multi-class open set recognition \cite{scheirer2014probability}.

% Open intent classification deals with the case that novel categories may appear during the inference. 

% The goal of open intent classification is to obtain a discriminator that can discriminator 
% Currently, when we do anomly detection 
% The open set classification is different from anomaly detection
% In anomly detection, we only need 

% The concept of recognizing new category is similar to the anomaly detection problem. The most anomaly detection problems mainly deal with generating a score to rank the known data and the unknown data in the inference. But they seldom discuss how to select a proper threshold to discriminate the nominal and anomaly data. A naive threshold will lead to bad performance as shown in \cite{zhang2021deep}

% Even if we perfectly rank the normal (known) data and the anomaly (unknown) data, bad performance is still possible if wrong. Thus the common metrics preferred is the PRC-ROC and the performance can be evaluated without a fixed threshold.
% The current work focuses on to build a boundary over the known data. 
% Comparing to the open-set classification, t
\input{imgs/draw_motivation}

To discriminate the known from the open samples during inference, it is necessary to create a clear classification boundary that separates the known from the open category.
However, the lack of open category samples during training makes this problem challenging.
Current research in this setting mainly focuses on two directions.
The first direction mainly estimates a tighter decision boundary between known classes to allow for the possibility of the open category. 
% Past works in this direction include the Local Outlier Factor (LOF)~\citep{breunig2000lof}, which is a manually designed metric that calculates the local density deviation of a data point from its neighbors.
Existing works in this direction include the Local Outlier Factor (LOF)~\citep{breunig2000lof, zhou2022knn, zeng2021modeling}, Deep Open Classification (DOC)~\citep{shu2017doc} and Adaptive Decision Boundary (ADB)~\citep{zhang2021deep}. LOF and ADB calibrate the decision boundary in the feature space while DOC does it in the probability space.

The second direction deals with learning a better feature representation to make the boundary-seeking problem easier. 
In this direction, DeepUnk~\citep{lin2019deep} and SEG~\citep{yan2020unknown} added constraints to the feature space, SCL~\citep{zeng2021modeling} and ~\citet{zhou2022knn} fine-tuned the feature extractor backbone with contrastive learning. \citet{zhan2021out} considered introducing open samples from other datasets as negative, and \citet{shu2021odist} generated samples with contradictory meanings using a large pretrained model. The latter two deliver large performance gains.

% Although external negative samples are introduced, huge improvements in~\citet{zhan2021out} and ~\citet{shu2021odist}
These improvements demonstrate the significance of negative samples in determining the boundary between the known and open categories. To accomplish the same in the absence of additional datasets or knowledge, we propose a novel negative-sample constraint and employ a gradient-based method to generate pseudo open category samples.
As shown in Figure~\ref{fig:motivation}(d), negative samples are \textit{adaptively} generated for each category to closely bound each category.
% instead of randomly distributing over the feature space~\citep{zhan2021out}.
% Based on this constrain, we then propose a gradient-based method to synthesize negative samples in the feature space.
% instead of generating real text.
% The \textit{virtual} negative samples in feature space are more effective and easier to obtain compared to the \textit{real} text negative samples from previous works.

%  In order to make the boundary between known and unknown tighter, we use a gradient-based method to synthesize samples near the positive one in the feature space. The negative samples are \textbf{adaptively} generated for each category, during which process, the data in the corresponding categories are used.
% we can control the distance between the positive and negative. We apply a gradient-based search algorithm to find negative samples near the positive ones.

Given the generated negative samples, we then empirically find that using auxiliary \ovr binary classifiers can better capture the boundary between the known and the open category, relative to a \((C+1)\)-way classifier~\citep{zhan2021out}, where all the open categories, possibly distributed in multiple modes or arbitrarily scattered over the feature space, are categorized into one class.

% However, the true distribution of the unknown samples might be multimodal and arbitrarily spread over the feature space~\citep{zhan2021out}.
% Hence, the generated examples are not effective and too easy for the model to learn a tight boundary.

%\citet{zhan2021out} utilized \((C+1)\)-way classifier and treated the negative sample introduced as a single open category. Unfortunately, the model may not be flexible enough to capture the possibly multimodal distribution of the unknown samples.
% comparing with using negative samples to construct a $(C+1)$ category classifier~\citep{zhan2021out}.
Specifically, we first learn a $C$-category classifier on known category data.
Then for each known category, we learn an auxiliary binary classifier, treating this category as positive and others as negative. 
% Then for each known category, we learn an auxiliary binary classifier on this category data and the corresponding adaptive negative samples. 
During inference, one sample is recognized as open if all the binary classifiers predict it as negative, thus not belonging to any known category

Our main contributions are summarized below:
%
% \begin{itemize}
%     \setlength\itemsep{-0.3em}
%     \item We propose a novel adaptive negative-sample-generation method for open-world classification problems without the need for external data or prior knowledge of the open categories. Moreover, negative samples can be added to existing methods and yield performance gains. 
%     \item We construct a new open-world classification workflow with auxiliary one-versus-rest binary classifiers, which learns better decision boundaries and requires no tuning (calibration) on the open category threshold.
%     \item We conduct extensive experiments to show that our approach significantly improves over previous state-of-the-art methods.
%     % \item We conduct extensive experiments on three benchmark datasets to show that our approach significantly improves over the previous state-of-the-art methods by a large margin.
% \end{itemize}
% previous version
\begin{itemize}
    \setlength\itemsep{-0.3em}
    \item We propose a novel adaptive negative-sample-generation method for open-world classification problems without the need for external data or prior knowledge of the open categories. Moreover, negative samples can be added to existing methods and yield performance gains. 
    % \item We construct a new open-world classification workflow with auxiliary one-versus-rest binary classifiers, which learns better decision boundaries and requires no tuning (calibration) on the open category threshold.
    \item We show that synthesized negative samples combined with auxiliary one-versus-rest binary classifiers facilitate learning better decision boundaries and requires no tuning (calibration) on the open category threshold.
    \item We conduct extensive experiments to show that our approach significantly improves over previous state-of-the-art methods.
\end{itemize}

%% file: imgs/draw_motivation.tex
\begin{figure}
% \begin{minipage}[t]{0.5\textwidth}
    % \vspace{1pt}
    % \hspace{1pt}
    \centering
    \includegraphics[width=0.46\textwidth]{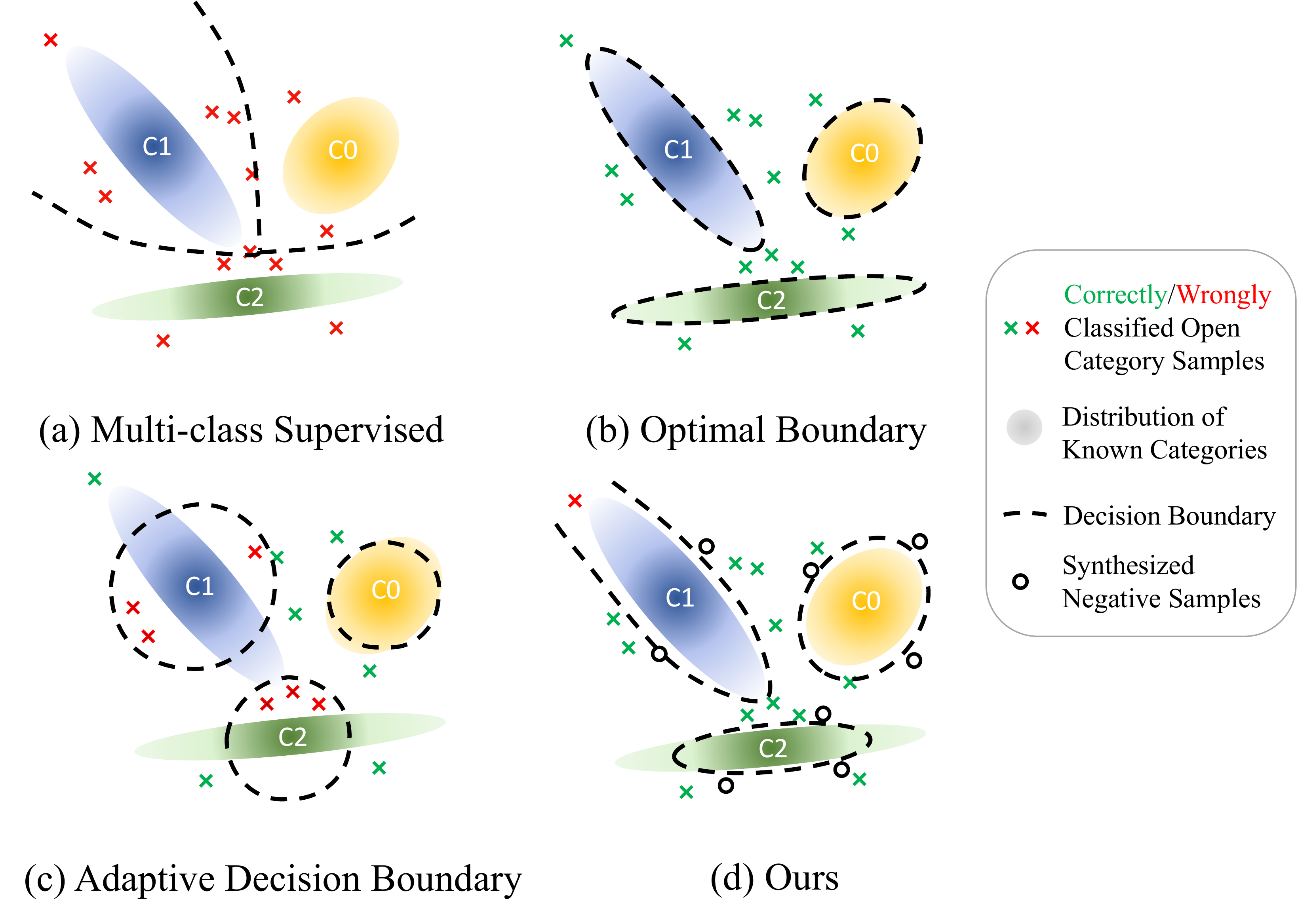}
    % \vspace{-8pt}
    % \caption{(Top) Conditional density. (Middle) Population density. (Bottom) Likelihood ratio.}\label{img:ll}
    % \vfill
%   \begin{minipage}[t]{1\textwidth}
    \caption{Illustration of previous methods and our proposed one. \(C0\), \(C1\), and \(C2\) are three categories. The boundary is used to discriminate known and open (unknown) categories. (a) Boundary learned by supervised learning. (b) Optimal decision boundary.
    % the one-vs-rest classifiers, which can well capture the boundary but need carefully selecting threshold.
    % It can well capture the boundary in the space constructed by the convex combination of samples, while ignoring the outlier ones. The threshold need to be adjusted carefully. 
    (c) ADB method which has a closed decision boundary, but may capture irrelevant points. (d) Proposed ANS method. 
    % \kkb{synthesized, not synthesis in legend}
    % (\textcolor{red}{(enlarge the triangle icon,change(b) looser(0.5) change (c), maybe add lof), Fig 1 add a OOD category?}
    }
    \label{fig:motivation}
%   \end{minipage}
% \end{minipage}
\vspace{-6pt}
\end{figure}

%% file: related.tex
\section{Related Work}
% We focus on work that only use 
% Though we introduce related works in two aspects, 
% \paragraph{Open Set Recognition: Boundary Calibration}
\textbf{Boundary Calibration}
The classical local outlier factor (LOF)~\citep{breunig2000lof} method is a custom metric that calculates the local density deviation of a data point from its neighbors.
% ~\citep{lin2019deep,yan2020unknown}.
% which uses the features from a multi-class model trained on known classes can be leveraged as an outlier detection metric to open class classification~\citep{lin2019deep,yan2020unknown}.  
% Given the features pretrained using multi-class classification loss on the known data, the classic Local Outlier Factor (LOF)~\citep{breunig2000lof} is commonly used as an off-the-shelf manually designed outlier detection metric~\citep{lin2019deep,yan2020unknown}.
% If the LOF score is under a threshold, the example is recognized as open. 
However, there is not a principled rule on how to choose the outlier detection threshold when using LOF. ~\citet{zeng2021modeling, zhou2022knn} added open category data into the validation set to estimate or grid-search the proper threshold.
So motivated, \citet{bendale2016towards} fit the output logits of the classifier to a Weibull distribution, but still use a validation set that contains the open category to select the confidence threshold.
Further, \citet{shu2017doc} employed one-versus-rest binary classifiers and then calculates the threshold over the confidence score space by fitting it to a Gaussian distribution.
This method is limited by the often inaccurate (uncalibrated) predictive confidence learned by the neural network~\citep{guo2017calibration}.
Adaptive decision boundary~\citep{zhang2021deep}, illustrated in Figure~\ref{fig:motivation}(c), was recently proposed to learn bounded spherical regions for known categories to contain known class samples.
Though this post-processing approach achieves state-of-the-art performance, it still suffers from the issue that the tight-bounded spheres may not exist or cannot be well-defined in representation space.
Due to the fact that high-dimensional data representations usually lie on a low-dimensional manifold~\citep{pless2009survey}, a sphere defined in a Euclidean space can be restrictive as a decision boundary.
Moreover, the issue can be more severe if certain categories follow multimodal or skewed distributions.
\input{imgs/draw_pipeline}

% \paragraph{Open Set Recognition: Feature Learning}
\noindent \textbf{Representation Learning}
DeepUnk~\citep{lin2019deep} trains the feature extractor with Large Margin Cosine Loss~\citep{wang2018cosface}. SEG~\citep{yan2020unknown} assumes that the known features follow the mixture Gaussian distribution. 
~\citet{zeng2021modeling} and ~\citet{zhou2022knn} applied supervised contrastive learning~\citep{chen2020simple} and further improve the representation quality by using $k$-nearest positive samples and negative samples collected from the memory buffer of MOCO~\citep{he2020momentum}. 
Getting a better representation trained with known category data only is complementary to our work, since a better pretrained backbone can further improve our results.
% These two representation models are learned directly from known category data.
%and later use LOF to detect unknown samples.
% After training on the known categories, LOF is employed to detect the unknown samples.
Recent works found that it may be problematic to learn features solely based on the known classes; thus, it is crucial to provide samples from unknown classes during training.
Specifically,
% \rhg{[ref]}.
\citet{zhan2021out} creates negative samples with mixup of training data and examples from an external dataset. \citet{shu2021odist} generates open class samples using BART~\citep{lewis2019bart} and external text entailment information.

%% file: imgs/draw_pipeline.tex
\begin{figure*}
\begin{minipage}[t]{\textwidth}
    \vspace{1pt}
    % \hspace{1pt}
    \centering
    \includegraphics[width=\textwidth]{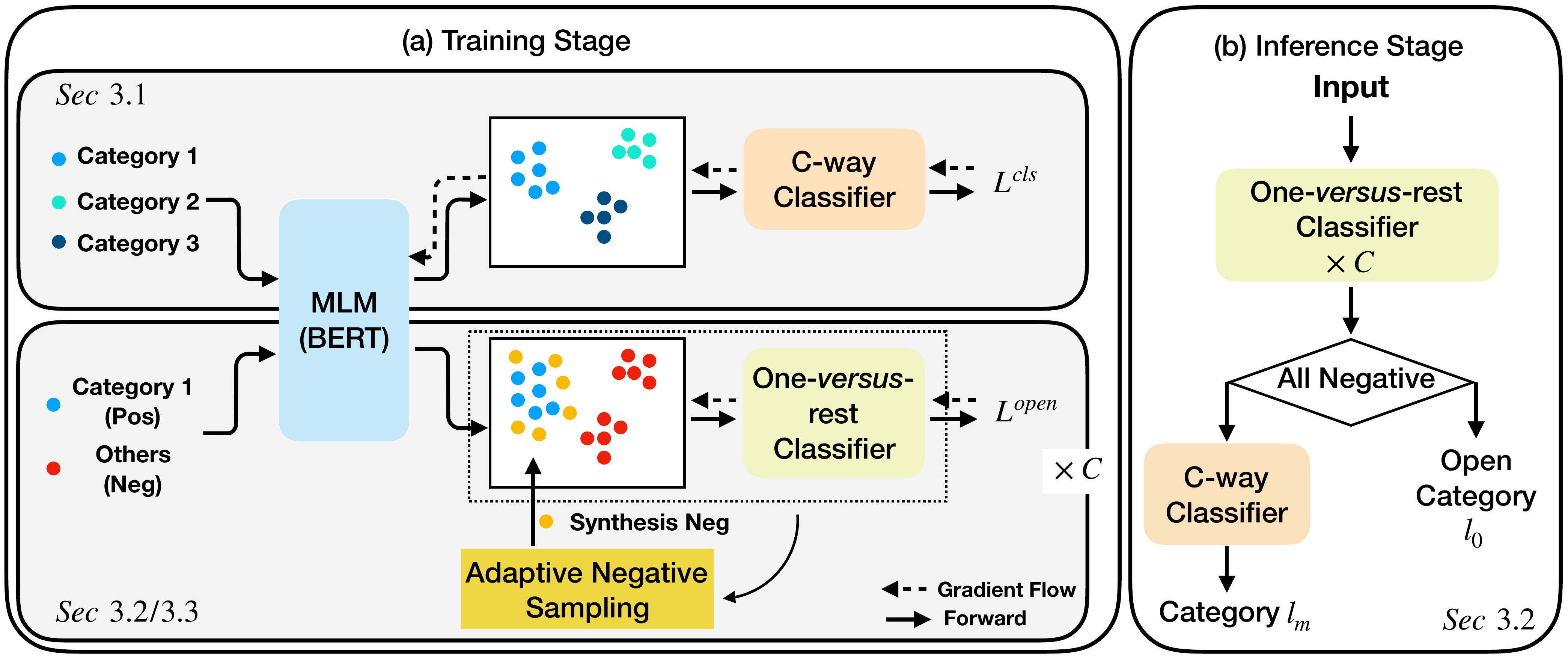}
    % \vspace{-8pt}
    \caption{Illustration of the proposed ANS algorithm. (a) The training stage is divided into two blocks. Top: known class classification described in Sec.~\ref{sec:subsec:known_cls}. Bottom: \ovr classification (Sec. ~\ref{sec:subsec:unknown_cls}) on Category 1 as an example. The negative samples are from two sources, namely, known from other classes (in red) and synthesized samples (in yellow).
    (b) Inference arm described at the end of 
    ~\Secref{sec:subsec:unknown_cls}.
    % \kkb{change down-right corner}
    % (b) Inference pipeline described in Sec.~\ref{sec:inference}.
    % (c) Illustration of our designed adaptive negative sampling process. Yellow dots are the generated samples.
    }
    \label{fig:pipeline}
    % \vfill
%   \begin{minipage}[t]{1\textwidth}
    % \caption{}\label{fig:pipeline}
%   \end{minipage}
\end{minipage}
% \vspace{-4mm}
\end{figure*}

%% file: methods.tex
\section{Methodology}
\paragraph{Problem Definition}
Suppose we are given a training dataset $\train =\{(\xv_1,y_1), (\xv_2,y_2), \ldots, $  $(\xv_N, y_N)\}$ consisting of $N$ examples, where $\xv_i$ is an input text sequence and $y_i$ its corresponding label, 
which belongs to a predefined set $\mathcal{L}=\{l_1, l_2, \ldots, l_C\}$ with $C$ categories, 
thus $ y_i \in \mathcal{L}, \forall i \in [N]$, where \([N] \coloneqq [1,\ldots,N]\). In this paper, we use $[\cdot]$ to represent index sequence.
In an open world classification setting, the goal is to learn a model which categorizes a test instance to either one of the predefined $C$ categories or as an open category.
In practice, the open category is denoted as a unique category $l_0$.
% Thus, the open world classification problem can be reformulated as a $(C+1)$ category classification problem.

% \subsection{Text feature representation}

\subsection{Known Category Classification}
\label{sec:subsec:known_cls}
%
% We first train a 
% We follow the setting in BERT and ADP.
% Follow the similar 
% Fine-tune a BERT and a classifier. 
% Following the setup in previous works~\citep{zhang2021deep,zhan2021out}, we use BERT~\citep{devlin2018bert} as the backbone network to extract features 
Following the setting suggested in~\citet{lin2019post,zhang2021deep}, 
% we first use BERT as our feature extractor $f_{\psi_{enc}}(\cdot)$ to get contextualized embedding of tokens \(h_{t_i}\) and then take the average of features $h_{t_i}$ of each token $t_i$ extracted from the BERT output layer as the sentence representation \(\hv\), \ie,
% we first use BERT as our feature extractor $f_{\psi_{enc}}(\cdot)$ and  contextualized embedding of each tokens \(h_{t_i}\) and then take the average of these features as the sentence representation \(\hv\), \ie,
we use BERT~\citep{devlin2018bert} as our feature extractor $f_{\psi_{enc}}(\cdot)$. For each input text sequence $\xv$, where $\xv$ is represented as a sequence of tokens $[t_1, t_2, \ldots, t_n]$, we take the average of features $z_{t_i}$ of each token $t_i$ extracted from the BERT output layer as the sentence representation \(\zv\).                                                                                                                                                                                           -                                              
% \ie,
% 
% \begin{align}
%     \zv &= \text{mean}(z_{t_1}, z_{t_2}, \ldots, z_{t_n} )  \\
%     & = \text{mean(BERT}(t_1, t_2, \ldots, t_n) \text{)}\in \sR^{768}.
%     \label{eq:feature_extractor}
%     \begin{aligned}
%     \end{aligned}
% \end{align}
% 
% where $f_{\psi_{enc}}(\cdot)$ is first adapted on the training data $\train$ to capture the characteristics of the known categories.
The training of $f_{\psi_{enc}}(\cdot)$ is formulated as a multi-category classification  problem by minimizing the cross-entropy loss $\Ls^{cls}$:
% The feature extractor $f$ is pre-trained using cross-entropy loss  \(L_{CLS}\) to do multiple category classification. 
% if one sample is recognized as ``known'' data during the inference, 
%
\begin{align}
		& \Ls^{cls}(\psi_{enc}, \psi_{cls})= \label{eq:xc} \\
		& \hspace{20mm}-\sum_{i\in[N]} \log\frac{\exp(f_{\psi_{cls}}^{y_{i}}(\zv_i))}{\sum_{j=1}^{C}\exp(f_{\psi_{cls}}^{j}(\zv_{i}))}, \notag
\end{align}
%
% where $N$ is the total number of training samples.
where $f_{\psi_{cls}}(\cdot)$ is a classifier that takes \(\zv\) as input and the output dimension is the number of known categories $C$. 
$f_{\psi_{cls}}^j(\zv)$ represents the output logit for the $j$-th category.
% The classifier is a fully connected neural network composed of one hidden layer with ReLU activation function.
% with the text feature $\hv$ as the input.
% The trainable parameters of the encoder and classifier are $\psi_{enc}$ and $\psi_{cls}$, respectively.
A well-trained feature extractor $f_{\psi_{enc}}(\cdot)$ and a high-quality classifier $f_{\psi_{cls}}(\cdot)$ can extract representative features of each category and ensure good performance on the classification of known categories during the inference stage.

% fully connected neural network 

% We pretrain the feature extractor with the multi-class cross-entropy loss. 

\subsection{Open Category Recognition}
\label{sec:subsec:unknown_cls}
Once the classifier for the known categories is available, the task is to recognize samples from the open category {\em versus} the ones from known categories.
As mentioned above, directly using the known category classifier $f_{\psi_{cls}}(\cdot)$ can result in poor performance~\citep{hendrycks2016baseline}, while using a $(C+1)$ category classifier setting is complicated due to the need to find proper samples to obtain a suitable decision boundary for the open category~\citep{scheirer2012toward, liang2017enhancing, shu2021odist}.
In this work, building upon ideas from one-class classification~\citep{scholkopf2001estimating, ruff2018deep} and one-{\em vs}-all support vector machines~\citep{rifkin2004defense}, we propose a \ovr framework via training simple binary classifiers for each predefined category.
Based on this framework, we then introduce an effective open-sample generation approach to train these classifiers in \Secref{sec:sub_ans}.
% In this subsection, we only focus on recognizing the unknown samples from known samples, without the attention to classify the known sample into a specific category, which could be accomplished in the classification step described in the previous paragraph.

%As we described in the introduction, our work used the one-versus-others framework. When the model is complex enough, the performance of the classifier is dependent on the quality of the positive and negative data. Because of this, We will first discuss how this framework works, and then introduce our method on generating adaptive negative samples.

% \paragraph{Auxiliary One-{\em versus}-rest Binary Classification}
%
We build an auxiliary \ovr binary classifier for each known category, and take $m$-th category as an example to illustrate.
Given a text sequence $\xv$, 
% we first obtain the corresponding input feature by utilizing the pre-trained feature extractor $f_{enc}$ and following the same aggregation step in Eq.~\ref{eq:feature_extractor} and feed it to the binary classifier. 
%
% we first obtain the corresponding input feature $\hv$ by utilizing the pre-trained feature extractor with parameter \(\theta^{enc}_m\), following the same aggregation step in Eq.~\ref{eq:feature_extractor}, and feed it to the binary classifier. The $m$-th class classifier is a learnable function $g_{\theta^{cls}_m}:\sR^d\rightarrow \sR$ such $g_{\theta^{cls}_m}(\xv)>0$ when the corresponding text $\xv$ belongs to the $m$-th class and vice versa.  In practical, the structure of each binary classifier is a 2-layer fully connected layer with ReLU activation function and Dropout. For convenience, we parameterise the whole classifier using \(\theta_m = (\theta^{enc}_m, \theta^{cls}_m)\).
%
we use the BERT pretrained with classification loss as the feature extractor $f_{\psi^{enc}}(\cdot)$ to extract features $\zv\in\sR^d$ to be fed to the binary classifiers, where $d$ is the dimension of the feature space.
Each category is provided with a binary classifier denoted as $g_{\theta^{cls}_m}(\zv):\sR^d\rightarrow \sR$, such that if $g_{\theta^{cls}_m}(\zv)>0$ then the input text $\xv$ belongs to the $m$-th category or vice versa belongs to any of the other categories.
We parameterize the entire binary classification framework for the $m$-th category as \(\theta_m = (\psi^{enc}, \theta^{cls}_m)\).
% We parameterize the entire binary classification framework for the $m$-th category as \(\theta_m = (\theta^{enc}, \theta^{cls}_m)\).
% During training, we fix the parameters of the feature extractor $\psi^{enc}$.
% Note that \(\zv\) could be either the contextualized token representations extracted from one layer of BERT, or the mean pooling of features from the last layer as shown in \Eqref{eq:feature_extractor}.
% Note that \(\zv\) could be either the contextualized token representations extracted from one layer of BERT, or the mean pooling of features from the last layer as shown in \Eqref{eq:feature_extractor}.
% (\(d = 768\)).
% In the first case, $d$ is the product of the sentence length and $768$ (the embedding size), its corresponding binary classifier consists of two modules, one is the last Transformer layer of the pretrained BERT. 
% This part and the feature extractor 
% the second module is a fully connected neural network with $256$ hidden units.
% \bc{same as freeze the parameters of first 11 transformer layers of BERT}
% \rhg{I do not understand the sentence}
% This structure follows ~\citet{zhang2021deep}.
% In the second case, $d = 768$ and the binary classifier is the same as the second module in the first case. 
% \rhg{only?}
% \bc{these two could be different, we want to differentiate these two}
% The $m$-th class classifier is a fully connected neural network with one hidden layer $
%
% For convenience, we parameterize the whole classifier using \(\theta_m = (\theta^{enc}, \theta^{cls}_m)\). 
% \rhg{why is this convenient?}
 
To learn each binary classifier $g_{\theta^{cls}_m}(\cdot)$ from training data $\train$, we first construct a \emph{positive} set
% $\{\xv_1^m, \xv_2^m, \ldots, \xv_{N_m}^m\}$
$\{\xv_1, \xv_2, \ldots, \xv_{N_m}\}$
using data points with label $l_m$ from $\train$ and a \emph{negative} set 
% $\{\hat{\xv}_1^m, \hat{\xv}_2^m, \ldots, \hat{\xv}_{N-N_m}^m\}$
$\{\hat{\xv}_1, \hat{\xv}_2, \ldots, \hat{\xv}_{N-N_m}\}$
by data points not in category $l_m$ but also from $\train$. The total number of samples within category \(m\) is \(N_m\), and \(N-N_m\) is the number of remaining samples in \(\train\). 
% Then we extract their features  $\{\zv_1^m, \zv_2^m, \ldots, \zv_n^m, \hat{\zv}_1^m, \hat{\zv}_2^m, \ldots, \hat{\zv}_n^m, \}$ respectively.
Each binary classifier is optimized by minimizing the binary cross-entropy loss function $\Ls^{rest}$:
\begin{align}
    % \label{eq:embed}
    % \frac{1}{N_m}\Big(
   \Ls^{rest}&(\theta^{cls}_m) =  \sum_{i\in [N_m]}\log(1+\exp(-g_{\theta_m}(\xv_i)))\nonumber \\
    + & \sum_{i\in [N - N_m]}\log(1+\exp(g_{\theta_m}(\hat{\xv}_i))).
    %\Big).
    \label{eq:binary}
\end{align}
%
%
% Utilizing the feature extractor $f_{enc}$ pretrained in the previous step, we obtain the input feature $\hv$ of the binary classifiers similarly via Eq.~\ref{eq:feature_extractor}.  
%
%  We build a one-versus others binary classifier for each category.
%
% We still use the BERT as the feature extractor and use the weight pretrained in the previous multi-class classification as the initialization. Following the same way as we did in Eq.~\ref{eq:feature_extractor}, we get the feature $\hv$. Then the dimension of $\hv$ is reduced to $1$ with a 2-layer fully connected layer with ReLU activation and Dropout, represented by $f_{b}^m$.  
%
% The model is trained with logistic regression loss. We select $n$ samples 
% For category $m$, $n$ samples \(\{\xv^m_i\}\) are drawn from the category $m$ as the positive data. $n$ negative data \(\{\hat{\xv}^m_i\}\) are drawn uniformly from all the data in the left known categories. The logistic regression target function is formed as below.
%
% \begin{align}
%     % \label{eq:embed}
%   \Ls^{rest}(\theta^{cls}_m) = & \frac{1}{N_m}\Big(\sum_{i\in [N_m]}\log(1+\exp(-g_{\theta_m}(\xv_i^m)))\nonumber \\
%     + & \sum_{i\in [N - N_m]}\log(1+\exp(g_{\theta_m}(\hat{\xv}_i^m)))\Big).
%     \label{eq:binary}
% \end{align}
%
% The parameters of \(mth\) binary classifier is \(\theta_m\)
%
During the inference phase, 
% a sample $\xv_i$ can be classified to category $l_m$, if \(g_{\theta_m}(\xv_i) > 0\). Furthermore, to discriminate whether a sample is from a known or open category, 
we have
\begin{align*}
\hat{y}=
\begin{cases}
	\text {open,} & \text{if } g_{\theta_m}(\xv_i) < 0, \forall m \in [C];\\
% 	\mathcal{C}; \\
	\text {known}, & \text{ otherwise}.
\end{cases}
% 	\end{equation}
\end{align*}
We assign a testing example to the open category $l_0$ if it is detected as unknown in all $C$ binary classifiers. Otherwise, we pass the example to the known classifier to categorize the known categories. The entire inference pipeline is illustrated in \Figref{fig:pipeline}(b).
\subsection{Adaptive Negative Sampling (ANS)}
\label{sec:sub_ans}
% \paragraph{Adaptive Negative Sampling (ANS)}
% The previous section described the binary classification setting. 
%
In practice, it is problematic to learn a good binary classifier with only the aforementioned negative samples from \(\train\). The sample space of the open category is complicated, considering that new-category samples could originate from different topics and sources, relative to the known classes.
So motivated, some existing methods introduce additional external data as negative samples.
% cannot be well-sampled by the known category training set \(\train\)~\citep{hendrycks2019using,zhan2021out}.
% Considering the distribution shift between the training and test dataset, the classifier will suffer from high False Positive Rate(FPR) in the inference stage~\citep{goyal2020drocc}.  
% \rhg{this does not make any sense, why are you suddenly talking about distribution shift?}
% \bc{topic of another paper on open set recognition, will modify it}

To alleviate the issue associated with the lack of real negative samples, we propose to synthesize negative samples \(\Tilde{\xv}\).
Considering that it is hard to create actual open-category text examples, we choose to draw \textit{virtual} text examples \(\Tilde{\zv}\) in the feature space~\citep{miyato2016adversarial,zhu2019freelb}.
% Note that it is much harder to create actual open category text examples compared to creating \textit{virtual} text examples \(\Tilde{\zv}\) in the feature space~\citep{miyato2016adversarial,zhu2019freelb}.
% \rhg{what do you mean by perturbed}
% Hence, we choose to draw samples from the feature space 
% \(\zv\) 
% instead of from the actual token space, considering that 
Compared with the token space of text, the feature space is typically smoother~\citep{bengio2013better}, which makes it convenient to calculate gradients~\citep{wang2021energy}.
% In our framework, the feature space $\zv$ can be freely chosen from the feature extractor, \eg the sentence representation $\hv$ or the contextualized token representations $(h_1, h_2, \ldots, h_n)$.  

% Though negative samples exists as described in the previous paragraph. It is hard to achieve a good classifier considering the distribution shift between the training and testing dataset. The lack of data from the unknown category leads to the loose boundary between the known data. Many unknown samples are contained in the positive space of one known category.

% In order to shrink this positive space, we target on synthesising negative samples $\Tilde{x}$. We choose to draw samples from the feature space \(\zv\) instead of the token space considering that the latent space is typically smoother~\citep{bengio2013better} and is easy to calculate the gradient~\citep{wang2021energy}.  Moreover, it is of great efficiency.

For all known samples in a category $l_m$, points that are away from these samples can be recognized as \emph{negative} to classifier \(g_{\theta_m^{cls}}(\cdot)\). The entire feature space \(\sR^d\) contains essentially an uncountable set of such points, among which we are mostly interested in those near the known samples.
% Such points can help us find the boundary of the known data.
% For each known sample \(\zv_i\), we can find a threshold $r$. Samples whose distance is larger than $r$ can be recognized as ``open''. Since the features are located on the low dimensional manifold, we can easily find such 
% So motivated, for each known sample \(\zv_i\), we can define a small ball with radius \(r\) around it.
% The union of points on the surface of ball can be seen as the boundary of these known samples when \(r\) is set properly.
% \rhg{sentence does not make sense, sphere of ball? are you taking about points inside, the surface of the ball or outside it? in any case, why is that useful. Also what do you mean by setting it properly.}
% typo surface
% An idea \(r\) should neither be too large to cover open samples, nor too small to generalize the known category samples in the testing set.
% \rhg{how do we know then if it is done properly?}
% With this boundary, any point in the feature space \(\sR^d\) that is outside of the union of these small balls can be considered as part of the open category~\citep{Madry2018TowardsDL}.
% Since there are uncountable many of such points, in practice, we are interested in points closest to any of the balls, from which we can understand as being closer to the boundary of known data.
% There are uncountable such points in the whole space, among which, if one point is closer to any ball, we could say that it is closer to the boundary of known data.
Consequently, capturing these points will be helpful to characterize the decision boundary.

% Here, 
To give a mathematical description of these points, we assume that data usually lies on a low-dimensional manifold in a high-dimensional space~\citep{pless2009survey}. 
% \sj{I don't think smooth manifold means isometric mapping between the manifold and $R^n$ unless one trained the model that way. In that sense, I think applying virtual adversarial training wrt L2 norm can be a good extension of this work. Anyway, I think we should soften this sentence.} 
The low-dimensional manifold can be viewed as a local-Euclidean space, thus we can use the Euclidean metric to measure distances locally for each known data $\zv_i$. 
% To define the distance, we assume that data lies on a low-dimensional manifold in a high-dimensional space~\citep{pless2009survey};
% The proposed method is based on a fundamental assumption. 
% In general, 
% and $(ii)$ a good synthetic open sample should be close to the predefined domain manifold, but not on the manifold.
% Given such open examples, the decision boundary can adapt to balance empirical and open space risk~\citep{bendale2015towards}.
% \rhg{empirical and open space risks are not defined}
% \bc{sorry, forget to delete it.}
    % \ie, maintaining high accuracy and reducing the FPR.
    % \item The synthesised negative samples should be close to the manifold. But it should not on the manifold.
%\end{itemize}
%
% Given these assumptions, the local manifolds can be treated as Euclidean spaces.
% The low dimensional manifold can be viewed as a local-Euclidean space, thus we can use the Euclidean metric to measure distances locally.
Under this assumption, the set of pseudo negative samples $\negset{i}(r)$ for \(\zv_i\), which we call \emph{adaptive synthesized open set}, can be described as follows,
% , for each known sample $\zv_i$. 
%
\begin{align}
\negset{i}(r)\eqdef \Big\{\Tilde{\zv}: & r\leq \| \tilde{\zv} - \zv_j \|_2 ;
        \label{eq:hyper-shell}
    \| \tilde{\zv} - \zv_i \|_2  
     \leq \gamma \cdot r, \notag \\
    % \nonumber\\
     &~\forall j: y_j = m \Big\}, 
\end{align}
where $r$ is the distance radius and $\gamma > 1$ is a hyperparameter.
Note that each known sample $\zv_i$ has an associated adaptive synthesized open set.
As defined above, this set is subject to two inequalities. 
The first keeps synthesized samples away from all known samples within category \(m\). 
The second implies that the synthesized samples should not be too far from the known samples. 
An intuitive geometric interpretation is that when \(j = i\), the space implied by these two constraints is a spherical shell with inner radius \(r\) and outer radius \(\gamma \cdot r\).

 To get the radius \(r\), we first calculate the covariance matrix \(\Sigma\) of \(\zv\) using known samples from category \(m\) and choose \(r\), s.t. \(r \le \sqrt{2\Tr(\Sigma)}\) and \(\gamma r \geq  \sqrt{2\Tr(\Sigma)}\). This is under the consideration that \(\sqrt{2\Tr(\Sigma)}\) is the average Euclidean distance between random samples drawn from a distribution with covariance matrix \(\Sigma\).
 
The estimation is supported by the following proposition,
\begin{proposition}
\label{prop}
% The average euclidean
The expectation of the euclidean distance between random points sampled from  distribution with covariance matrix \(\Sigma\) is smaller than \(\sqrt{2\Tr(\Sigma)}\), \ie
% Samples \(\zv_i \sim \) 
% \sqrt{\|\zv_i - \zv_j\|^2}

% \begin{align}
%     \E_{\xv, \yv\sim\mathcal{N}(\muv, \Sigma)}
%      \sqrt{\|\xv - \yv\|^2}     \leq \sqrt{2\Tr{\Sigma}}
%     \end{align}
% \end{proposition}

\begin{align}
    \E_{\xv, \yv\sim\mathcal{D}(\muv, \Sigma)}
     \sqrt{\|\xv - \yv\|^2}     \leq \sqrt{2\Tr{\Sigma}}
    \end{align}
\end{proposition}

The proof can be found in supplementary. In our experiments, we fix \(\gamma=2\) and \(r = 8\). The choice of \(r\) is relevant to the covariance matrix of the features in representation space. The detailed justification for our selection is provided in Appendix~\ref{app:sec:radius}. Ablation studies (\Figref{fig:exp_radius}) show that model performance is not very sensitive to the chosen \(r\).

\paragraph{Binary Classification with ANS}
According to \Eqref{eq:hyper-shell}, each sample from a known category $m$ contributes an adaptive synthesized set of open samples \(\negset{i}(r)\). 
The classifier \(g_{\theta_m^{cls}}(\cdot)\) is expected to discriminate them as negative.
% The classifier \(g_{\theta_m^{CLS}}\) is expected to discriminate them out from the samples in category \(m\).
The corresponding objective function is the binary cross-entropy loss,
% the logistic regression loss that only has negnative samples as the input.
% with only the negative part, aiming to optimize classification parameters that categorize synthesized open samples as negative for binary classifier \(g_{\theta_m}^{cls}\),
% \bc{becky q1: how to express we sample many points from \(\negset{i}(r)\), here exists one problem}
%
\begin{align}
     \Ls^{syn}(\theta^{cls}_m) =
    %  \label{eq:feature} 
     %\frac{1}{N_m} \Big(
     \sum_{i\in [N_m]}  \log(1+\exp(g_{\theta^{cls}_m}(\Tilde{\zv}_i))),
     %\Big),
     \notag
\end{align}
where \(\Tilde{\zv}_i\) is sampled from \( \negset{i}(r)\) and $N_m$ is the total number of known samples with category $l_m$.
% , and $\negset{i}(r)$ is the synthesised open set built for each point $\zv_i$ with category $m$.
However, there exist uncountably many points in \(\negset{i}(r)\). Randomly sampling one example from $\negset{i}(r)$ is not effective.
% However, this will result in a huge \(\negset{i}\) that will make it difficult to iterate over every single point during training.
% due to the huge space of \(\negset{i}\), we cannot iterate every point. 
% Among pointless samples in \(\negset{i}\), 
Alternatively, we choose the most representative ones that are hard for the classifier to classify it as \emph{negative}. 
% This concept is also similar to select the 
%
Consistent with this intuition, the \(\max(\cdot)\) operator is added to select the most challenging synthetic open sample distributed in $\negset{i}(r)$.
\begin{align}
    & \Ls^{syn}(\theta^{cls}_m) = \label{eq:feature} \\
    % \frac{1}{N_m} \Big(
    &\hspace{4mm}  \sum_{i\in [N_m]} \max \limits_{\tilde{\zv}_i\in\negset{i}(r)}  \log(1+\exp(g_{\theta^{cls}_m}(\Tilde{\zv}_i))).
    %\Big),
    \notag
\end{align}
%
% \bc{the summation means that we iterate all the samples in dataset with category m. for each sample, we select one open sample in N(r)},
%
 
% \rhg{mention that the hardest is the closest to the ball boundary}
% \bc{becky q2: it should not be the closest to the ball boundary?}
% Thus, \eqref{eq:feature} aims to obtain classification parameters that categorize synthesized open samples as negative for binary classifier \(g_{\theta_m}^{cls}\).
% \rhg{as negative by all binary classifiers, right? if so, please fix}
% \bc{fixed}

Finally, the complete loss accounting for open recognition is summarized as:
% \begin{align}
%     \Ls^{open} = \sum_{i\in [C]}(\Ls^{rest} + \lambda \Ls^{syn}), 
% \end{align}
\begin{align}
\label{eq:open_loss}
    \Ls^{open} = \Ls^{rest} + \lambda \Ls^{syn}, 
\end{align}
where $\lambda$ is a regularization hyperparameter.

Directly minimizing the objective function in \Eqref{eq:open_loss} subject to the constraint in \Eqref{eq:hyper-shell} is challenging. In the experiments, we adopt the projected gradient descend-ascend technique~\citep{goyal2020drocc} to solve this problem.
\begin{table*}[h]
\begin{center}
\begin{sc}
  \scalebox{0.85}{
\begin{tabular}{@{}cccccccc@{}}
\toprule
\multirow{2}{*}{\%} & \multicolumn{1}{l}{} & \multicolumn{2}{c}{BANKING}   & \multicolumn{2}{c}{CLINC}   & \multicolumn{2}{c}{StackOverflow} \\ \cmidrule(l){2-8} 
                    & Methods              & Accuracy       & F1-score       & Accuracy       & F1-score       & Accuracy        & F1-score        \\ \midrule
\multirow{7}{*}{25} & MSP                  & 43.67          & 50.09          & 47.02          & 47.62          & 28.67           & 37.85           \\
                    & DOC                  & 56.99          & 58.03          & 74.97          & 66.37          & 42.74           & 47.73           \\
                    & OpenMax              & 49.94          & 54.14          & 68.50          & 61.99          & 40.28           & 45.98           \\
                    & DeepUnk              & 64.21          & 61.36          & 81.43          & 71.16          & 47.84           & 52.05           \\
                    & ADB                  & 78.85          & 71.62          & 87.59          & 77.19          & 86.72           & 80.83           \\
                    & SelfSup*             & 74.11          & 69.93          & 88.44          & 80.73          & 68.74           & 65.64           \\
                    & Ours                 & \textbf{83.93} & \textbf{76.15} & \textbf{92.64} & \textbf{84.81} & \textbf{90.88}  & \textbf{84.52}  \\ \midrule
\multirow{7}{*}{50} & MSP                  & 59.73          & 71.18          & 62.96          & 70.41          & 52.42           & 63.01           \\
                    & DOC                  & 64.81          & 73.12          & 77.16          & 78.26          & 52.53           & 62.84           \\
                    & OpenMax              & 65.31          & 74.24          & 80.11          & 80.56          & 60.35           & 68.18           \\
                    & DeepUnk              & 72.73          & 77.53          & 83.35          & 82.16          & 58.98           & 68.01           \\
                    & ADB                  & 78.86          & 80.90          & 86.54          & 85.05          & \textbf{86.40}  & {85.83}  \\
                    & SelfSup*             & 72.69          & 79.21          & 88.33          & 86.67          & 75.08           & 78.55           \\
                    & Ours                 & \textbf{81.97} & \textbf{83.29} & \textbf{90.23} & \textbf{88.01} & 86.08           & \textbf{85.90}           \\ \midrule
\multirow{7}{*}{75} & MSP                  & 75.89          & 83.60          & 74.07          & 82.38          & 72.17           & 77.95           \\
                    & DOC                  & 76.77          & 83.34          & 78.73          & 83.59          & 68.91           & 75.06           \\
                    & OpenMax              & 77.45          & 84.07          & 76.80          & 73.16          & 74.42           & 79.78           \\
                    & DeepUnk              & 78.52          & 84.31          & 83.71          & 86.23          & 72.33           & 78.28           \\
                    & ADB                  & 81.08          & 85.96          & 86.32          & 88.53          & 82.78           & 85.99           \\
                    & SelfSup*             & 81.07          & \textbf{86.98} & 88.08          & 89.43          & 81.71           & 85.85           \\
                    & Ours                 & \textbf{82.49} & 86.92          & \textbf{88.96} & \textbf{89.97} & \textbf{84.40}  & \textbf{87.49}  \\ \bottomrule
\end{tabular}
}
\end{sc}
\end{center}
\caption{
% \sj{the best system is marked wrong for Banking 77 at 75\%} 
Results of open world classification on three datasets with different known class proportions. 
% The scores of baseline methods are from the~\citep{zhang2021deep} and ~\citep{zhan2021out}.
* indicates the use of extra datasets during the training. The first five results of the baseline model are from ~\citet{zhang2021deep}. The results for SelfSup are from ~\citet{zhan2021out}.
% \footnotemark
% \footnote{The first six baseline model results are from ~\citet{zhang2021deep}.
% The results of SelfSup are from the corresponding paper. }
}
\label{table:main}
\end{table*}
% \paragraph{Projected Gradient Descend-Ascend}
% \noindent\textbf{Projected Gradient Descend-Ascent}
\paragraph{Projected Gradient Descend-Ascent}
\input{algs/ans}
% To solve this saddle point problem, 
We use gradient descent to minimize the open recognition loss $\Ls^{open}$ and gradient ascent to find the hardest synthetic negative samples $\tilde{\zv}$. The detailed steps are summarized in Algorithm~\ref{alg:ans}. 
% The pipeline is summarized in the Algorithm~\ref{alg:ans}.
% As illustrated in Figure~\ref{fig:pipeline}(c), the sample $\Tilde{\zv}'$ obtained from direct gradient ascend might be out of the constraint area $\negset{i}(r)$. 
As illustrated in Figure~\ref{fig:pipeline}(c), the sample $\Tilde{\zv_i}' = \Tilde{\zv_i} + \epsilon$ directly derived from gradient ascent (line 11 of Algorithm~\ref{alg:ans}) might be out of the constraint area $\negset{i}(r)$. 
We then project to the closest $\tilde{\zv_i}$ within the constraint such that $\Tilde{\zv_i} = \arg\min_\uv\|\Tilde{\zv_i}' - \uv\|^2, \forall \ \  \uv\in \negset{i}(r)$ ~\citep{boyd2004convex}.
Unfortunately, direct search within $\negset{i}(r)$ defined in \Eqref{eq:hyper-shell} requires complex computation over entire training data $\train$. 
Based on our assumption that the training samples lie on a low-dimensional manifold and the empirical observation that \(\Tilde{\zv}_i'\) is always closest to the corresponding positive point \(\zv_i\) relative to other positive points, \(\negset{i}(r)\) can be further relaxed to the sphere shell around sample \(\zv_i\): \( \negset{i}(r)= \{\Tilde{\zv}: r\leq \|  \tilde{\zv} - \zv_i \|_2 \leq \gamma \cdot r\}\). 
We can then directly find the synthetic negative sample via a projection along the radius direction, \ie, \(\Tilde{\zv_i} = \Tilde{\zv_i}' + \alpha \frac{\Tilde{\zv_i}' - \zv_i}{\|\Tilde{\zv_i}' - \zv_i\|}\), where \(\alpha\) is adjusted to guarantee \(\Tilde{\zv_i} \in \negset{i}(r)\):
\begin{align}
        \alpha= 
\begin{cases}
    1,& \text{if }r\leq \|  \tilde{\zv}_i - \zv_i \|_2 \leq \gamma \cdot r \\
    \frac{r\gamma}{\|  \tilde{\zv}_i - \zv_i \|_2} ,  &  \text{if }\gamma \cdot r \leq \|  \tilde{\zv}_i - \zv_i \|_2 \\
        \frac{r}{\|  \tilde{\zv}_i - \zv_i \|_2} ,  &  \text{if }  \|  \tilde{\zv}_i - \zv_i \|_2 \leq r \notag
\end{cases}
\label{eq:alpha}
\end{align}
In the future, we would like to consider relaxing these constraints by only considering the nearest \(k\) points instead of all the points within a category.

%% file: algs/ans.tex
\begin{algorithm}[t]
\footnotesize
\caption{
% \sj{shouldn't line 9 get outside of the loop?} 
\footnotesize Adaptive Negative Sampling.}
\label{alg:ans}
\begin{algorithmic}[1]
\State {\bfseries Input:} \footnotesize{Training data \(\train= \xv_1, \xv_2, \cdots, \xv_n, \hat{\xv}_1, \cdots, \hat{\xv}_n\)}. \footnotesize{Parameters of current binary classifier \(\theta_m\).} 
\State{\bfseries Hyper-Parameters:} \footnotesize{Radius $r$, step-size $\eta$, number of gradient steps $k$ }
% \State{\bfseries Representation learning steps:}
% \State{\For B = 1, \cdots, n_0}
\For{Batch number \(B = 1, \cdots, n_0\)}
    \State{\(X_B(\hat{X}_B)\): Collect a batch of positive (negative) samples.}
    \State{Calculate loss \(\Ls_1 = \Ls^{real}(X_B, \hat{X}_B)\) using Eq.~\ref{eq:binary}}
    \State{Calculate the feature $\zv_B$ over the positive samples.}
    \State{\bfseries Adaptive Negative Sampling:}
    % \State{\hspace*{\algorithmicindent} 
    \State{sample $\epsilonv \sim \sN(0, 4\cdot diag(\Sigma_\zv))$}
    \For{\(i = 1, \cdots, k\)}
    % get \(\zv_B' = \zv_B + \epsilonv\)}
    \State{Calculate loss \(\ell(\zv_B + \epsilonv)\) using Eq.~\ref{eq:feature}}
    % \State{Calculate loss \(\ell(\epsilonv) = \Ls^{syn}_{\theta_m'}(\zv_B')\) using Eq.~\ref{eq:feature}}
    \State{$\epsilonv = \epsilonv + \eta \frac{\nabla_\epsilonv \ell(\zv_B + \epsilonv)}{\| \nabla_\epsilonv \ell(\zv_B + \epsilonv) \|}$} \Comment{Gradient Ascend}
    \EndFor
    \State{Calculate \(\alpha\) using Eq.\ref{eq:alpha}}
    \State{$\epsilonv = \frac{\alpha}{\|\epsilonv\|}\cdot \epsilonv$  
    % \Comment{}
    % $\alpha= r\cdot \mathbbm{1}[\|\epsilonv\|\leq r]+ \|\epsilonv\|\cdot \mathbbm{1}[r\leq \|\epsilonv\|\leq \gamma \cdot r] + \gamma \cdot r\cdot \mathbbm{1}[\|\epsilonv\|\geq \gamma \cdot r]$
    } 
    \State{Calculate loss \(\Ls_2 = \Ls^{syn}_{\theta'_m}(\zv_B + \epsilonv)\) using Eq.~\ref{eq:feature}}
    \State{\(\theta_m = \theta_m - \nabla_{\theta_m}(\Ls_1 + \Ls_2)\)} \Comment{Gradient Descend}
    % \hspace*{60pt}
\EndFor
% \State{\bfseries ANS steps:}
\end{algorithmic}
\end{algorithm}

%% file: experiments.tex
\section{Experiments}
\def\num#1{\numx#1}\def\numx#1e#2{{#1}\mathrm{e}{#2}}

We conduct experiments on three datasets: Banking, CLINC and Stackoverflow. 
Details and examples of the datasets are found in Appendix \ref{app:sec:dataset_sum}. 
%  The hyperparameter of the experiment design is also included in the Appendix. 

% \noindent \textbf{Task Design}
\paragraph{Task Design}
We apply the same settings as ~\citet{shu2017doc} and ~\citet{ lin2019deep}. For each dataset, we sample \(25\%\), \(50\%\), \(75\%\) categories randomly and treat them as the  \emph{known category}.
Any other categories out of the known categories are grouped into the \emph{open category}.
In the training and validation set, only samples within the known category are kept.
All the samples in the testing set are retained, and the label of samples belonging to open categories is set to \(l_0\).
Importantly, samples from the open category are never exposed to the model in the training and validation process.

% \noindent\textbf{Evaluation Metrics}
\paragraph{Evaluation Metrics}
\label{sec:exp:para:eval}
The model needs to identify the samples with the open category, as well as classify the known samples correctly.
Following ~\citet{shu2017doc, zhang2021deep}, we use accuracy and macro \fscore as our evaluation metrics.
The accuracy measures the overall performance, considering that open-world classification can be treated as a $C+1$ classification problem.
\fscore is a binary classification metric mainly used for evaluating the performance of open category recognition. The F1-score reported in this paper is the mean of the macro F1-score per category (including the open category), where the positive category is the corresponding one and negatives are all the other categories.
F1-known is the average over F1-score of all known categories. F1-open is the F1-score of the open category. 
% More details about evaluation metric can be found in Appendix~\ref{app:sec:evalutaion}. 
% F1-score reported in this paper is the mean of macro F1-score per category (including the open), where the positive category is the corresponding one and negatives are all the other categories.
% \fscore-known is the average of macro F1-score of all known categories.  \fscore-open is the macro F1-score of the open category. 
% We also calculate the macro \(F1\) score over the known and unknown categories and put it in the appendix.
% \input{tabels/main_results}

% \noindent\textbf{Experimental setting}
\paragraph{Experimental setting}
% \subsection{}
% \paragraph{Multi-class Classification}
We use the BERT-base-uncased model to initialize the feature extractor \(f_{\phi_{enc}}\)
% For the 
and freeze the first ten layers of BERT during training.
% as in ~\citep{zhang2021deep} 
% of multi-category classification.
Note that all results are the mean of ten trials with different random seeds. Other experimental details are included in~\Appenref{sec:app:exp_setting}.
% \rhg{either number al the appendices or none.}
%  As a reminder, \textbf{OOD category only shows up in testing}.

% \paragraph{Auxiliary Binary Classifiers}
% We use the BERT pretrained in the multi-class classification as the initialization. The learning rate setting keeps the same as above. 

% The latent space \(\zv\) of CLINC and Banking dataset is the output of the second last BERT layer. 
% \(\zv\) for StackOverflow is the pooling of the features of the last BERT layer \(\zv \in \sR^{768}\), considering the fact that Stackoverflow dataset is simpler than the others. 
% About the parameters related to adaptive negative sampling, we set $\lambda=0.5$ considering the confidence of generated samples is lower than the known negatives. 
% The choice of hyperparameter in ANS follows the suggestions in ~\citep{goyal2020drocc}: gradient ascent learning rate \(\eta = \num{1e-1}\), ascent steps \(k = 5\). 
% The ratio of positive/known negative/synthesized negative is 1:1:1. The weight \(\lambda\) in \Eqref{eq:open_loss} is \(0.5\).
% The radius $r$ is $64$ for the feature using the output of the second last layer of BERT, $8$ for the features using the last layer with pooling. 
% We make this choice to let the square root of feature dimensions locates in the range \([r, \gamma r]\). 
% The ratio of positive/known negative/synthesized negative is 1:2:1 for 75\% setting and 1:1:1 for others.

\subsection{Results}

\input{tabels/discussion_negative}
\input{imgs/draw_radius}
Table~\ref{table:main} 
% \footnotetext{The first six baseline model results are from ~\citet{zhang2021deep}. The results of SelfSup are from the corresponding paper.} 
compares our approach with previous state-of-the-art methods using accuracy and \fscore. 
% The accuracy is the calculated on \(C+1\) categories, including the open category. F1 score is the average of Macro-F1 score for each category. 
%  Our implementation is based on the code base from  
Our implementation is based on~\citet{zhang-etal-2021-textoir}.
% \paragraph{baselines}
% The results are collected f
The baselines include threshold finding methods,
MSP~\citep{hendrycks2016baseline}, 
DOC~\citep{shu2017doc},
OpenMax~\citep{bendale2015towards}, 
ADB~\citep{zhang2021deep}; and
feature learning methods, 
DeepUnk~\citep{lin2019deep};  
and negative data generation method SelfSup~\citep{zhan2021out}. 
We did not include results from ODIST~\citep{shu2021odist} because their method relies on MNLI-pretrained BART, which is not currently public and their model performance drops dramatically if not coupled with ADB.
% The results of SelfSup are from the corresponding paper.}
Note that SelfSup uses additional datasets, without which the accuracy on 50\% CLINC drops from 88.33 to 83.12.
% , \fscore drops from 86.67 to 82.62.
% , we could reimplement it using the same codebase because of the lack of extra data. 
% \input{tabels/discussion_negative}

% \input{tabels/discussion_negative}

Our approach performs better than most previous methods, even better than the method using additional datasets, with the greatest improvement on CLINC. 
% Among all the baselines, 
% ADB~\citet{zhang2021deep} performs the best on BANKING and Stackoverflow. 
This is in accordance with SelfSup~\citep{zhan2021out}, which also benefits the most on CLINC by adding negative samples.
% when compared with ADB, the strongest baseline. 
This implies that our synthesized negative samples are of high quality and could possibly be used as extra datasets in other methods.

The average performance gain in these three datasets decreases as the known category ratio increases, \ie, compared to the strongest baseline ADB, our accuracy improvements in the three datasets are \(5.08, 3.11, 1.42\) under the setting of \(25\%, 50\%, 75\%\). 
With more known categories available, the more diverse the known negative samples will be, allowing the model to better capture the boundaries of the positive known categories while reducing the impact of synthetic samples.
% With more known categories, the negative known samples are more diverse, thus the boundary of the positive known category can be better captured, and the impact of the synthesized sample reduces. 

The comparison with baselines on \fscore-open and \fscore-known can be found in ~\Appenref{app:table:main}.
% \kkb{format}
% Our method improves the most on CLINC,
% Among all the three dataset, we find that the largest boost is on CLINC. 
% Considering that SelfSup is also a method benefits from negative samples, 
% This is in accordance with SelfSup, a method benefits the most from adding negative samples. 
% greater if the known category is less, \ie, in comparasion with ADB, the improvement on CLINC is \(2.76, 1.94, 1.77\) under the setting of \(25\%, 50\%, 75\%\). 
% This is reasonable since the boundary of one category could by constraint by more negatives sampled from other categories and the effectiveness of synthetic negatives decreases. 
% This phenomenon is also in accordance with the tendency that the gaps between all baselines decrease as ratio of known category increases.

% \input{imgs/draw_radius}
\input{tabels/ablation_neg}
\input{imgs/draw_tsne}

\subsection{Discussion}
% \input{tabels/discussion_negative}
% \paragraph{One-versus-rest Structure}
\paragraph{Synthesized Negative is Beneficial for a Variety of Structures}
% Previous experiments show 
% As we mentioned in the introduction, our work differentiates from the most previous work in two parts, one is the negative synthesize mechanism 

To investigate the contribution of the synthesized samples and the structure of \ovr, we performed experiments adding the synthesized samples to two well-known baselines, MSP~\citep{hendrycks2016baseline} and ADB~\citep{zhang2021deep} as shown in \Tableref{table:syn_negative}. 
Specifically, the C-way classifier in MSP and ADB is replaced by a ($C+1$)-way classifier, with an extra head for the synthesized negative samples. 
See~\Appenref{app:sec:ans} for details.
% about their training and inference processes. 

We observe that performance increases on all baselines with synthesized negative samples. 
The synthesized samples behave like data augmentation, leading to better representation of positive samples.
%  Firstly, the synthesized samples improve each baseline. 
% we have several findings.
% MSP and \ovr 
% ADB works on the feature space.
% The synthesized negative samples could act as data augmentation for ADB, leading to better representations of the positive sample. 
% The performance increases.
% MSP and \ovr both work in the probability space and employ \(0.5\) as the threshold to discriminate between the known and open categories. 
%  They both performs bad without synthesized samples.
% , the performance of MSP and \ovr are both bad.

Further, synthesized samples benefit \ovr the most.
The difference, we believe, stems from the model's flexibility on boundary learning. 
The open category may contain sentences with various themes, making it difficult for a single head of a ($C+1$)-way classifier to catch them all.
This \ovr flexibility comes at the cost of more classifier parameters. However, compared to the huge feature extractor BERT, the number of additional parameters is relatively small.
% To be specific, the number of the classifier parameter is about \(3.4\), \(6.8\), \(10.0\) times on Banking with \(25\%, 50\%, 75\%\) known categories, compared with SelfSup. 
% To further analysis the contribution of the synthesized negative samples and the \ovr framework. 
% If necessary, for example, there are thousands of tasks, distillation techniques can be used to create a smaller model.
Distillation techniques can be used to build a smaller model if necessary, for instance, where there are thousands of known categories.

% \noindent \textbf{Adaptive Negatives Samples Generation}
\paragraph{Adaptive Negatives Samples Generation}
Our adaptive negative sample generation consists of three modules (a) adding Gaussian noise to the original samples (line \(8\) in Algorithm~\ref{alg:ans}). (b) gradient ascent (line \(10\sim11\)) (c) projection (line \(13\sim14\)). 
We add each module to the baseline in turn to study their importance.
% The first and third step provides us negatives and the second step improve the quality of generated ones. 
% We design experiments to show the contribution of each part. 
% In Table~\ref{table:discussion_negative}, . 
% Each part is accumulated step-by-step.
The baseline experiment uses the vanilla \ovr framework described in \Secref{sec:subsec:unknown_cls}, without the use of synthesized negative samples. 
Experiments are conducted on CLINC as shown in \Tableref{table:discussion_negative}.

The following describes our findings from each experiment:
(\(i\)) 
Adding samples with noise as negative alleviates the overconfidence problem of the classifier and improves the results significantly. The noise level needs to be designed  carefully since small noise blurs the distinction between known and open, while large noise is ineffective.

(\(ii\)) Constraining synthesized samples to $\negset{}(r)$ improves performance by keeping synthesized samples from being too close or too far away from positive known samples.

(\(iii\)) Adding a gradient ascent step further enhances performance.
The improvement over the previous step is marginal.
Our hypothesis is that the calculated gradient could be noisy, since the noise we add is isotropic and may be inconsistent with (outside of) the manifold of the original data. 
\paragraph{Radius \(r\) Analysis}
% Adaptive negative sample generation process introduces hyper-parameters \ie \(\eta, r, \gamma, k \).
% These are parameters are related; a large ascent rate $r$ with a small ascent step $k$ could behave the same as a small ascent rate with a large ascent step. 
In the adaptive negative sample generation process,  the radius \(r\) and multiplier \(\gamma\) are two hyperparameters that determine the upper and lower bounds of the distance between the synthesized sample and the known sample.
To investigate the impact of radius, we fix $\gamma$ to 2 and increase the  $r$ from \(1\) to \(256\). Note that \(8\) is our default setting. 
% \todo{emphasize}

As illustrated in \Figref{fig:exp_radius}, 
the performance gradually drops when the radius $r$ increases, because the influence of the synthesized negative examples reduces as the distance between them and the positive samples grows.
% Because the distance between the synthesized negative samples and the positive ones is larger, their effect on boundary learning decreases. 
When the radius $r$ decreases, the classifier may be more likely to incorrectly categorize positive as negative because the synthesized negative samples are closer to the known positives, resulting in a decrease in accuracy and F1 score on the banking and CLINC datasets.
% . Classifier could be more likely to wrongly classify positive into negative, thus the accuracy and F1-score drop on dataset Banking and CLINC. 
However, the performance on Stackoverflow improves. We hypothesize that there is a better data-adaptive way to estimate the radius \(r\) to improve the performance even further, for example, using \(k\) nearest neighbor instead of all the data in a category.
% specifically for Stackoverflow. 
We leave this as interesting future direction. 

In summary, we observe that the performance is affected by the radius, but comparable results can be obtained for a wide value range. They are all better than the vanilla \ovr baseline, which lacks the generated negative samples. 
The accuracy of baselines on Banking, CLINC and Stackoverflow is \(58.09\), \(71.80\) and \(64.58\), respectively. 
% Full results can be found in Appendix. 
% \bc{future work: part 1}

% Secondly, ADB is a method that works on feature space. It can also benefit from the synthesized negative samples 
% In this paragraph, we are trying to figure out the gain from negative samples.

% \bc{future work: part 2}

% \Tableref{table:discussion_negative}
%
% \noindent
% \textbf{Visualization}
\paragraph{Visualization}
\Figref{fig:tsne} shows the $t$-SNE representation of the features extracted from the second hidden layer of \ovr classifier \(g_{m}^{cls}\).
% (Please enlarge for a better view),
% We choose the feature of the second hidden layer.
Randomly chosen three known categories, each corresponds to a \ovr classifier, yield three figures.
The known positive/negative samples (blue) are clustered because the features are extracted from a pretrained $C$-way classifier \Secref{sec:subsec:known_cls}.
%  , thus the known samples clustered.
Open samples (pink) are scattered, some of which overlap with the known positives (see the middle figure). Our synthesized negatives work as expected; they are close to the known positives and bridge the gap between the positive and other known categories.

%% file: tabels/discussion_negative.tex
\begin{table}[]
\begin{center}
\begin{sc}
   \scalebox{0.68}{
\begin{tabular}{llcccc}
\hline
\%                  & Methods                  & Acc            & F1             & F1-open        & F1-known       \\ \hline
\multirow{4}{*}{25} & Baseline ($\lambda = 0$) & 57.05          & 53.12          & 63.83          & 52.84          \\
                    & \(+\) Gaussian Noise     & 90.37          & 82.09          & 93.75          & 81.78          \\
                    & \(+\) Projection         & 92.02          & 83.99          & 94.91          & 83.71          \\
                    & \(+\) Ascend (Ours)      & {92.32} & {84.34} & {95.11} & {84.05} \\ \cline{2-6} 
\multirow{4}{*}{50} & Baseline ($\lambda = 0$) & 64.60          & 71.65          & 60.28          & 71.80          \\
                    & \(+\) Gaussian Noise     & 88.01          & 86.90          & 89.82          & 86.86          \\
                    & \(+\) Projection         & 90.22          & 88.18          & 92.01          & 88.12          \\
                    & \(+\) Ascend (Ours)      & {90.23} & {88.22} & {92.02} & {88.17} \\ \cline{2-6} 
\multirow{4}{*}{75} & Baseline ($\lambda = 0$) & 76.17          & 83.63          & 63.23          & 83.81          \\
                    & \(+\) Gaussian Noise     & 88.67          & 90.45          & 86.71          & 90.48          \\
                    & \(+\) Projection         & 88.89          & 89.95          & 87.52          & 89.97          \\
                    & \(+\) Ascend (Ours)      & {88.96} & {89.97} & {87.62} & {90.00} \\ \hline
\end{tabular}
}
\end{sc}
\end{center}
\caption{Ablation study on the negative samples generation. The results are conducted on CLINC with known classes proportion 25\%, 50\% and 75\%. Baseline represents the experiment with no synthesized negative samples under the \ovr framework. }
\label{table:discussion_negative}
% \vspace{-3mm}
\end{table}

%% file: imgs/draw_radius.tex
\begin{figure}[t]
% \begin{minipage}[t]{0.5\textwidth}
    % \vspace{1pt}
    % \hspace{1pt}
    \centering
    \includegraphics[width=0.5\textwidth]{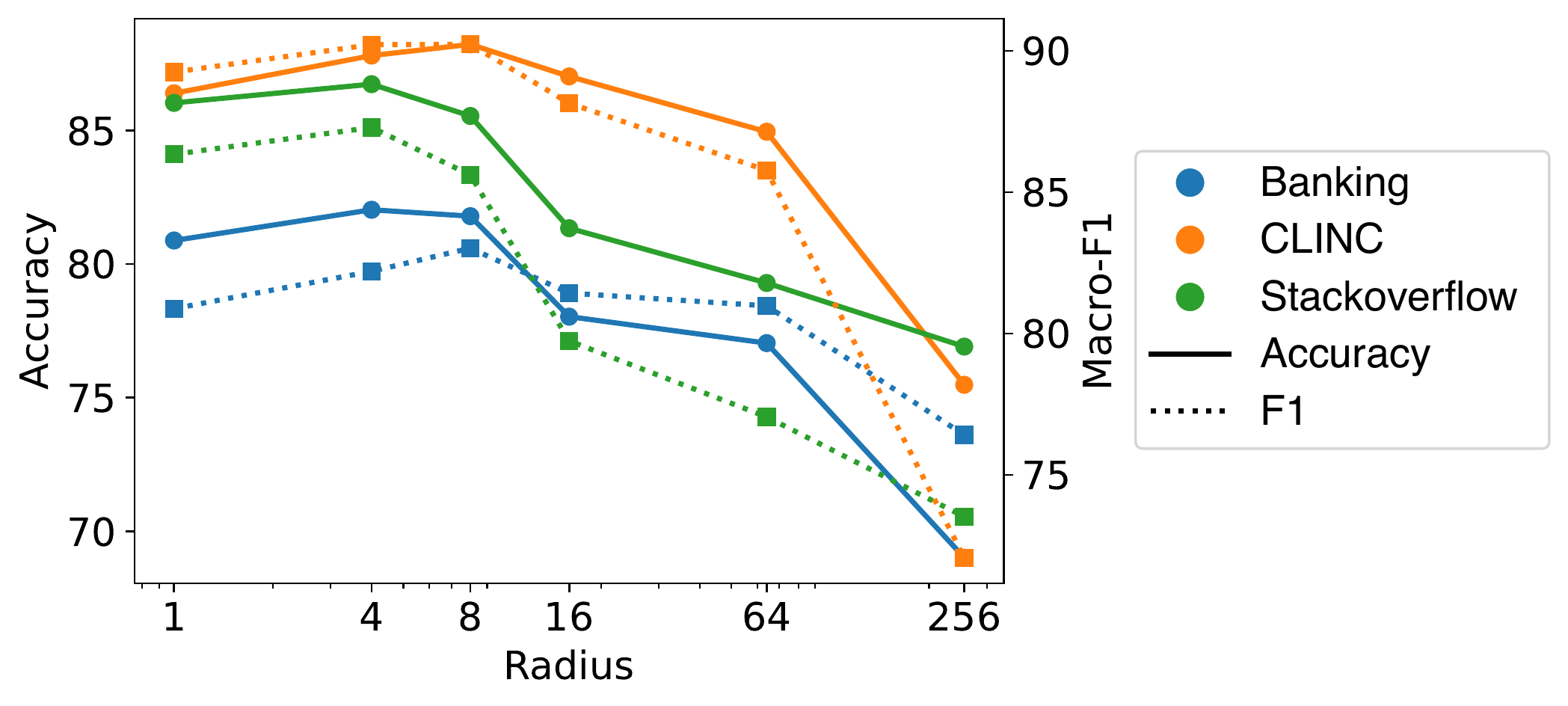}
    % \vspace{-8pt}
    % \caption{(Top) Conditional density. (Middle) Population density. (Bottom) Likelihood ratio.}\label{img:ll}
    % \vfill
%   \begin{minipage}[t]{1\textwidth}
    \caption{Ablation study on the radius \(r\) used in adaptive negative sampling process under the setting with 50\% known categories.
    }
    \label{fig:exp_radius}
%   \end{minipage}
% \end{minipage}
% \vspace{-3mm}
\end{figure}

%% file: tabels/ablation_neg.tex
\begin{table}
\begin{center}
\begin{sc}
   \scalebox{0.7}{
\begin{tabular}{@{}cccccc@{}}
\toprule
Methods                                                                 & Add Neg.       & Acc   & F1    & F1-open & F1-known \\ \midrule
\multirow{2}{*}{MSP}                                                    &                & 44.46 & 51.95 & 43.22   & 52.41    \\
                                                                        & \(\checkmark\) & 57.07 & 58.92 & 61.40   & 58.79    \\ \midrule
\multirow{2}{*}{ADB}                                                    &                & 78.39 & 71.53 & 84.18   & 70.86    \\
                                                                        & \(\checkmark\) & 79.71 & 73.01 & 85.18   & 72.12    \\ \midrule
\multirow{2}{*}{\begin{tabular}[c]{@{}c@{}}One-vs\\ -Rest\end{tabular}} &                & 51.33 & 41.40 & 50.83   & 43.54    \\
                                        %   & 80.11 & 73.35 & 85.57   & 72.71 
                                        % & \(\checkmark\) & 80.27 & 73.42 & 85.71   & 72.89    \\
                                                                                & \(\checkmark\) & 80.11 & 73.35 & 85.57   & 72.71    \\
                                        \bottomrule
\end{tabular}
}
\end{sc}
\end{center}
\caption{Performance comparison when the synthesized negative data are added to different baselines. The experiments are conducted on Banking with 25\% known categories.}
\label{table:syn_negative}
\end{table}

%% file: imgs/draw_tsne.tex
\begin{figure*}[ht]
\minipage{0.32\textwidth}
  \includegraphics[width=\linewidth]{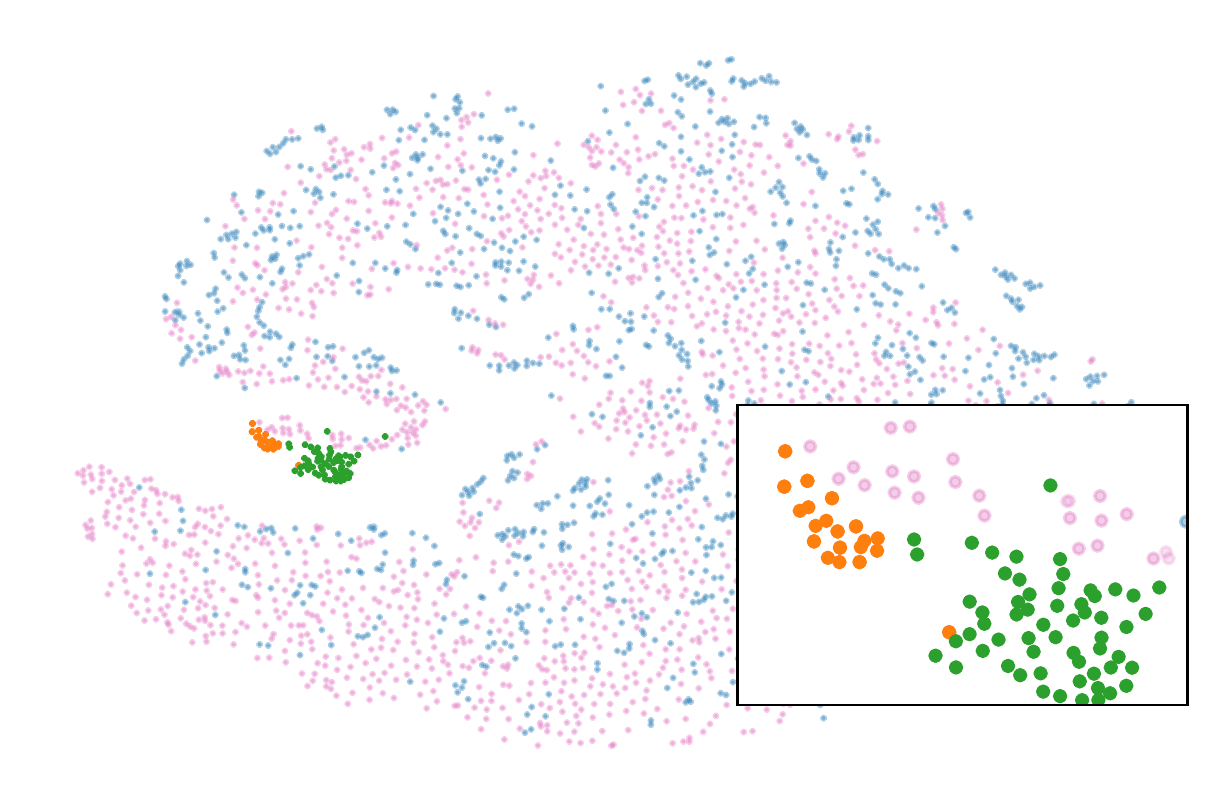}
%   \caption{A really Awesome Image}\label{fig:awesome_image1}
\endminipage\hfill
\minipage{0.32\textwidth}
  \includegraphics[width=\linewidth]{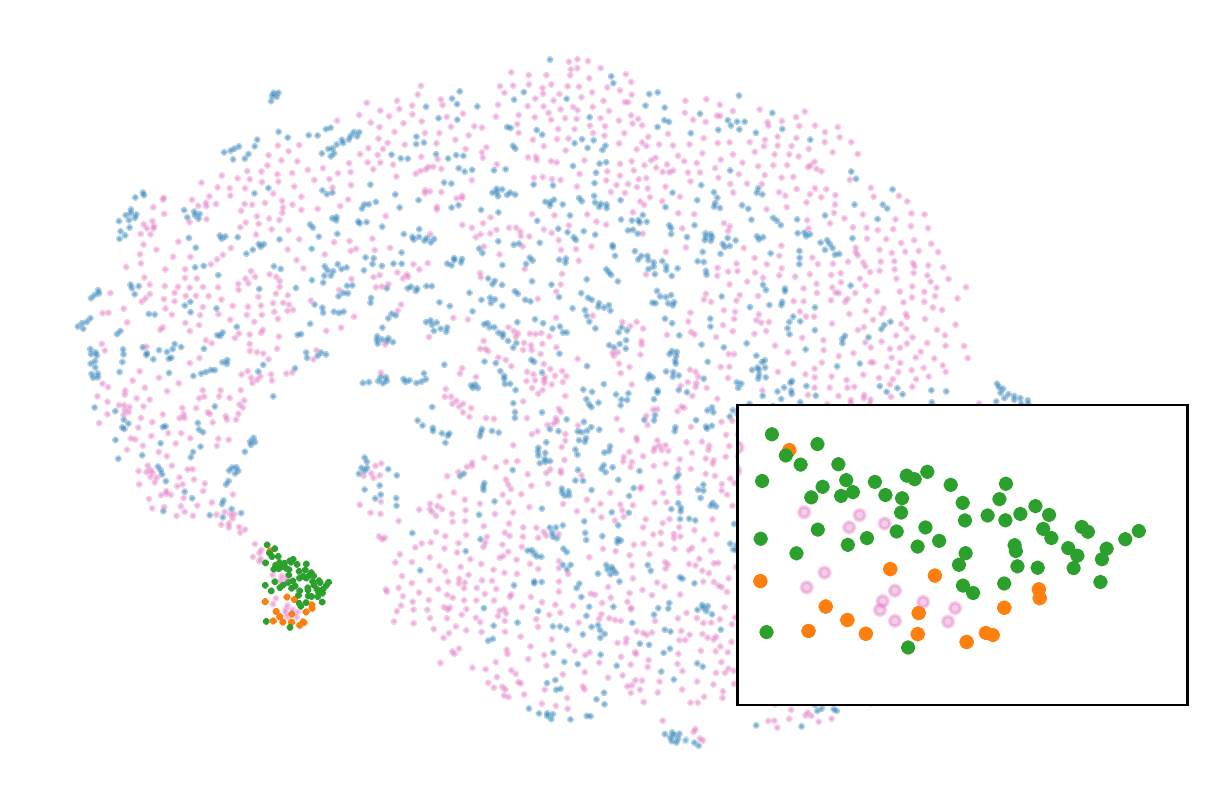}
%   \caption{A really Awesome Image}\label{fig:awesome_image2}
\endminipage\hfill 
% \vfill
% \minipage{0.3\textwidth}%
%   \includegraphics[width=\linewidth]{imgs/oos50_9.pdf}
% %   \caption{A really Awesome Image}\label{fig:awesome_image3}
% \endminipage\hfill 
\minipage{0.32\textwidth}%
  \includegraphics[width=\linewidth]{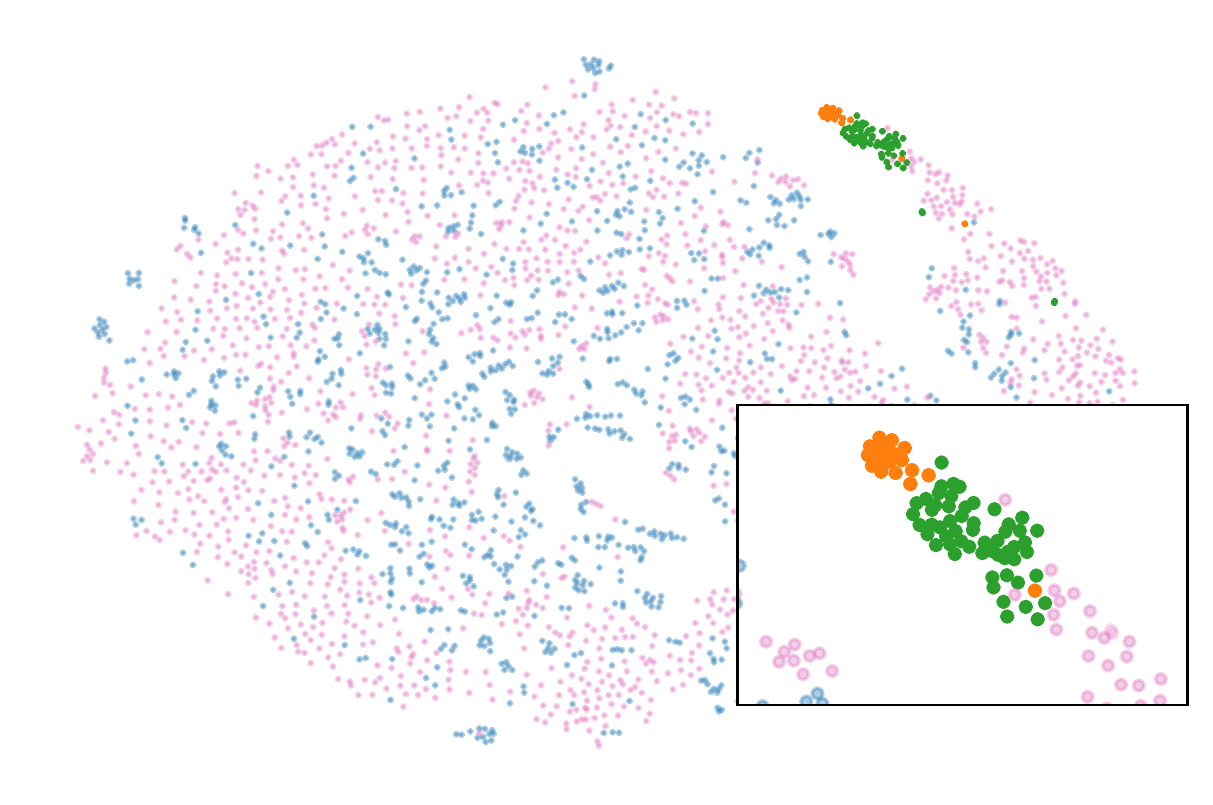}
%   \caption{A really Awesome Image}\label{fig:awesome_image3}
\endminipage\\
\vfill
\minipage{\textwidth}%
\centering
  \includegraphics[width=0.8\linewidth]{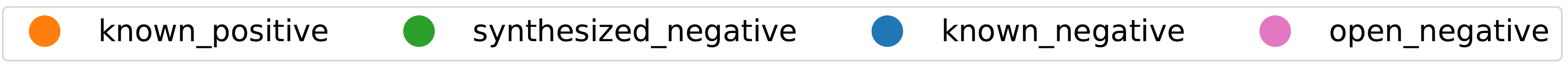}
%   \caption{A really Awesome Image}\label{fig:awesome_image3}
\endminipage
\caption{
$t$-SNE plots of the feature extracted from the testing set of CLINC with 50\% known categories.
% TSNE plots of the feature extracted from the testing set of CLINC with 50\% known categories.
% Better viewed in color.
Each panel corresponds to a \ovr classifier $g_{\theta_m^{cls}}$ with different known category \(m\) acting as the positive. 
Each panel's lower right corner has a square that enlarges the known positive data region to show the effectiveness of the synthesized negative samples.
% ``Known\_positive(negative)'' are clustered because they exist in the training set and are trained by classification loss.
% selects one known category as the positive and others as known negative. 
% The features are extracted from the testing set of CLINC with 50\% known categories. 
(Best viewed in color).
% The model is trained on CLINIC dataset with \(50\%\) known categories. 
% Nine squares correspond to nine categories randomly picked from known categories. The blue/orange dots represent the known/synthesized samples separately.
}
% \vspace{-3mm}
\label{fig:tsne}
\end{figure*}

%% file: conclusion.tex
\section{Conclusions}
% In our methods, we generate 
We have introduced ANS, a pseudo open category sample generation approach for open-world classification problems.
% The samples are adaptiv generated 
% The previous methods can be improved by adding 
The generation process is free from extra datasets or prior knowledge. The synthesized samples are effective for improving existing methods.
Combined with \ovr framework, significant improvements are observed on three benchmarks. 
The gradient-based negative sample generation proposed can be applied to other NLP tasks such as out-of-scope discovery, which we leave as future work.

%% file: limitations.tex
\section{Limitations}
First, both in terms of the number of training samples and the length of sentences, all the datasets that we employ in this study are relatively small and short. It may not scale well for long texts. 
Second, further work should be done on the model's ability to handle increasingly complicated data. The training samples from three benchmark datasets might be too sample; many input sentences even include the category descriptions, as shown in Tabel~\ref{app:table:demo}.

%% file: acknowledgement.tex
\section{Acknowledgement}
This research was supported by DARPA, DOE, NIH, ONR and NSF.

%% file: supplementary.tex
\section{Appendix}
\label{sec:appendix}

% Second, the capability of the model on more complex data deserves further study as well. As shown in Tabel~\ref{app:table:demo}, the training sample may even contain the category descriptions. 

\subsection{Ethical Consideration}
The topics of the three datasets we use in this paper are relatively simple, covering only the information needed for classification (check \Tableref{app:table:demo}). The category labels are either everyday intentions or technical terms in computer science. There are no potentially sensitive topics or contents that we are aware of. All three datasets have been published and are included in our appendix.

\renewcommand\thetable{\thesection.\arabic{table}}    
\setcounter{table}{0}    
% \paragraph{Summary}

\subsection{Related work: Adversary Augmentation}
The gradient descend-ascend technique has been used successfully in adversarial attacks~\citep{Madry2018TowardsDL,zeng2021modeling}; however, it differs from ours in terms of motivation and loss formulation.

~\citet{Madry2018TowardsDL} sought for samples that were similar to the training sample but had a substantial loss given the paired label. The addition of generated samples during training may strengthen the model's resistance to adversarial attacks by avoiding inputs that are indistinguishable from genuine data but improperly categorised. The associated optimization formula is
\begin{align}
    \min_\theta \ \E_{\xv,y\sim D} [\max_{\delta \in S} l(\theta, \xv+ \delta, y)],
\end{align}

where $y, \xv$ are the training data, and \(l\) could be any classification model parameterized by \(\theta\). \(S\) is the an adversarial perturbation \(l_\infty\)  ball.

% However, the purpose of 
% They tried to find 
%

Our work targets on shrinking the decision boundary. We need to treat the samples with positive labels in a specific region \(\negset{}(r)\)(defined in \Eqref{eq:hyper-shell}) as negative, \ie
\begin{align}
    \min_\theta \ \E_{\zv \sim D} [\max_{\delta \in \negset{}(r)} l(\theta, \zv + \delta, -1)],
\end{align}

where \(\zv\) is the positive sample from dataset \(D\), \(l\) is a binary classifier with parameter \(\delta\). This equation behaves the same as~\Eqref{eq:feature}.

\subsection{Explanation on Radius \(r\)}
\label{app:sec:radius}
% \sj{given our appeal to the local property of the manifold, isn't it more natural to consider the radius of the local point cluster of a point rather than the entire class cluster?} 
% \bc{yes, we could consider the k-nearest neighbors}
% We choose the radius \(r\) smaller than the average Euclidean distance between two random variables drawn from distribution with covariance \(\Sigma\),
% where \(\Sigma\) is the covariance matrix of the latent space \(\zv\) extracted from samples in the positive category. 
% In the future, we might consider calculating 

%  To get the radius \(r\), we first calculate the covariance matrix \(\Sigma\) of \(\zv\) using known samples from category \(m\) and choose \(r\), s.t. \(r \le \sqrt{2\Tr(\Sigma)}\) and \(\gamma r \geq  \sqrt{2\Tr(\Sigma)}\). This is under the consideration that \(\sqrt{2\Tr(\Sigma)}\) is the average Euclidean distance between random samples drawn from a distribution with covariance matrix \(\Sigma\).
 
% The estimation is supported by the following proposition,
\begin{proposition}
\label{prop}
% The average euclidean
The expectation of the euclidean distance between random points sampled from  distribution with covariance matrix \(\Sigma\) is smaller than \(\sqrt{2\Tr(\Sigma)}\), \ie
% Samples \(\zv_i \sim \) 
% \sqrt{\|\zv_i - \zv_j\|^2}

% \begin{align}
%     \E_{\xv, \yv\sim\mathcal{N}(\muv, \Sigma)}
%      \sqrt{\|\xv - \yv\|^2}     \leq \sqrt{2\Tr{\Sigma}}
%     \end{align}
% \end{proposition}

\begin{align}
    \E_{\xv, \yv\sim\mathcal{D}(\muv, \Sigma)}
     \sqrt{\|\xv - \yv\|^2}     \leq \sqrt{2\Tr{\Sigma}}
    \end{align}
\end{proposition}

\label{app:sec:proof}
% \sj{this derivation doesn't seem right. Please double-check the proof.} 
Proof: Given that we are measuring the distance between samples drawn from the same distribution, we could subtract a constant value from both variables and assume that the distribution's expectation is zero.
If \(\xv\) and \(\yv\) are random variables independently sampled from  distribution with covariance matrix \(\Sigma\) and zero mean, we could have:
\begin{align*}
    \E(\|\xv - \yv\|^2_2) 
    & = \sum_i \E(\xv_i^2 - 2\xv_i\yv_i + \yv_i^2)\\
    & = 2\sum_i (\E(\xv_i^2) - \E(\xv_i)\E(\yv_i)) \\
    & = 2\sum_i (\E(\xv_i^2) \\
    % & + \E(\xv_j)\E(\xv_j)^T -
    % 2\E(\xv_i)\E(\xv_j)^T \\
    & = 2 \Tr(\Sigma)
\end{align*}
% \begin{lemma}\label{prop:distance}
% % The average distance between random sampled from distribution
% if \(\xv\) is sampled from normal distribution \(\mathcal{N}(\mu, \Sigma)\), then:
% \begin{align}
%     \E(\|\xv_i - \xv_j\|^2) = 2  \textit{tr}(\Sigma)
% \end{align}
% \end{lemma}
% add it later.

% \begin{corollary}
For a random variable $Z$, Jensen's inequality gives us
\begin{align}
    \E(\sqrt{Z}) \leq \sqrt{\E[Z]}
\end{align}

The combination of the two equations above proves the proposition,
\begin{align*}
    \E(\|\xv - \yv\|_2)\leq  \sqrt{2\Tr(\Sigma)}
\end{align*}
% \end{corollary}

% The equality supports when \(\E[Z]\)
% \end{lemma}

% \begin{proposition}

% \end{proposition}

% Assume that there exists one Gaussian distribution 
% \(\sigma_i = std(z_i^{j})\)
% \begin{align}
%       \sqrt{2 \sum(\sigma_i^2 )}
% \end{align}

% In our setting, we assume that \(\Sigma = \sigma I_d\), thus the average euclidean distance between samples from \(\mathcal{N}(\mu, \sigma I_d)\) is 
% \begin{align}
%     \E(\sqrt{\|\zv_i - \zv_j\|^2}) \leq \sqrt{2d}\sigma
% \end{align},
% considering the fact that \(\text{tr}(\sigma^2 I_d) = \sigma^2 d\).

In experiment, we choose the mean of the last layer of BERT as the latent representation $\zv \in \R^{768}$. 
% We make a simple statistics of element-wise standard deviation of \(\zv\) per category and they mainly fall into an area between 8 to 12. 
% get the value \(0.6, 0.6, 0.5\) for dataset BANKING, CLINIC and Stackoverflow separately. Their corresponding value is \(16.63, 16.63, 13.86\). To ensure that \(\sigma \sqrt{d}\in[r , \gamma r] \), we set \(r = 16, 16, 8\) for each of them. 
When calculating the trace, only the variance of each dimension, are required.  
On three datasets, the predicted distance per category falls primarily into \([8, 12]\), we fix \(r=8\) for all the experiments. 

% people may raise concern here, 
For each positive point, we could sample several adaptive negative samples. The distance between the synthesized negative and the chosen positive is determined by \(r\). Meanwhile, we can also calculate the distance between the synthesized negative and other positive known samples.

We find that even when the radius is set to be less than the average distance, the synthesized negative samples have a much greater distance to other known points. In theory, known samples are on a low-dimensional manifold, whereas synthesized points are in a high-dimensional space, and the probability of sampled points falling into the manifold is zero. We calculated the distance between the synthesized sample and other known samples in the same category empirically, and discovered that their distances are nearly twice the average distance.

A further ablation study on different options of the radius can be found in the main context, where we have comparisons over different radius, \ie, \(r \in \{1, 4, 8, 16, 64\}\).

% \begin{proposition}\label{prop:kl}
%     \(p(\xv)=\sum_t\!m^{(t)}q^{(t)}(\xv)\) is optimal for minimizing the expected KL divergence
%     % \begin{small}
%     % \begin{align}\label{eq:min_kl}
%     %     \min_{p(\xv)\in\mathcal{P}}\mathbb{E}[\operatorname{KL}(q^{(t)}|p)] =\min_{p(\xv)\in\mathcal{P}}\sum_t\mathbb{E}_{q^{(t)}(\xv)} \left[ m^{(t)} \log \frac{q^{(t)}(\xv)}{p(\xv)} \right].\vspace{-3pt}
%     % \end{align}
%     % \end{small}
%     \resizebox{.95\linewidth}{!}{
%   \begin{minipage}{\linewidth}
%   \begin{align}
% \min_{p(\xv)\in\mathcal{P}}\mathbb{E}[\operatorname{KL}(q^{(t)}|p)] =\min_{p(\xv)\in\mathcal{P}}\sum_t\mathbb{E}_{q^{(t)}(\xv)} \left[ m^{(t)} \log \frac{q^{(t)}(\xv)}{p(\xv)} \right]
% \end{align}
%   \end{minipage}
% }
% \end{proposition}
\input{tabels/app_dataset_domo}
 \input{tabels/dataset}

\subsection{Dataset}
All three datasets are in English. The label distributions of each dataset are balanced.
\label{app:sec:dataset_sum}
\textbf{Banking} \citep{casanueva2020efficient} is a data set for intent detection in the banking domain with 77 fine-grained categories. It contains 13,083 samples in total. 
We follow the original splitting and sampling, \ie, 9,003, 1,000, and 3,080 for training, validation and testing respectively. 

% \ie, 15,000, 3,000, and 4,500 for training, validation and testing respectively. 

\noindent \textbf{CLINC}~\citep{larson2019evaluation} is an intent classification dataset originally designed for out-of-scope detection. 
 It owns 22,500 in-scope samples and 1,200 out-of-scope samples that existed only in the testing phase. 
 The divisions on the in-scope ones follow the original training, validation and testing splitting, \ie, 15,000, 3,000 and 5,700. 
 
\noindent \textbf{StackOverflow}~\citep{xu2015short} consists of 20 categories of technical question titles. 
 We utilize the prepossessed version with 12,000 for training, 2,000 for validation and 6,000 for testing.
 
The statistics of the datasets are summarized in \Tableref{tab:dataset_stats}.
We also provide raw data samples of each dataset in \Tableref{app:table:demo} for an intuitive understanding of the open world recognition task.

% Check 
% \subsection{Evaluation Metric}
% \label{app:sec:evalutaion}
% F1-score reported in this paper is the mean of macro F1-score per category (including the open), where the positive category is the corresponding one and negatives are all the other categories.
% \fscore-known is the average of macro F1-score of all known categories.  \fscore-open is the macro F1-score of the open category. 
% 
\subsection{Experimental Details}
\label{sec:app:exp_setting}

All experiments are executed on a single NVIDIA Titan Xp GPU with 12,196M memory.
All the experiments are done on NVIDIA X 
\paragraph{Known Category Classification}
The classifier \(f_{\psi_{cls}}\) is a fully connected neural network composed of one hidden layer with ReLU activation function. The hidden size is 768. 

The learning rate of the transformer part and non-transformer part are \(\num{5e-5}\) and \(\num{1e-4}\) respectively. 
The total number of training epochs is 100. Learning rate decay and early stops are applied. 

Training on this part takes about 10-20 minutes, depending on when the early stopping is triggered.

\paragraph{\ovr Binary Classification}
During the training of \ovr structure, we fix the parameters of the feature extractor $\psi^{enc}$.

For the \ovr module, the feature \(\zv\) is chosen as the mean of the output of the BERT model's last layer. The classifier is a fully connected three-layer neural network with ReLU as the activation function. The numbers of hidden neurons are (256,64), respectively. Dropout is added per hidden layer. The learning rate of the classifier is \(\num{1e-3}\) for Banking and CLINC, \(\num{3e-4}\) for Stackoverflow. The total number of epochs for each classifier is \(\min(C, 20)\) to avoid overfitting. \(\gamma\) is \(0.5\).

Currently, we train each \ovr classifier individually, and this takes roughly a minute per head. As a result, the total time grows linearly with the number of known categories. Parallel training of multiple heads can increase efficiency if necessary.

\paragraph{Model Size}
The parameters of the BERT backbone model and the ovr classifier are approximately 109 million and 0.2 million, respectively, implying that the maximum number of parameters from \ovr is only about one-fifth of BERT (75\% CLINC). 

\paragraph{Reproducibility Checklist: hyper-parameter search}
We didn't include results from the validation set considering there is a huge gap between the current validation set and test set; test set contains open category samples while the validation set does not.

It is difficult to study the hyper-parameter setting because we lack an effective validation set with open category samples and the testing set is unavailable during training. To solve this, we construct a "pseudo dataset" by selecting a subset as "sub-known" from all known categories and treating others as "sub-open". Taking 50\% CLINC as an example, we take a quarter of the known category as "sub-known" and the others as "sub-open". We discover that the rules we developed using these synthesized datasets can be transferred to formal experiments. We choose the proper hyper-parameter according to \fscore.

The hyperparameters we manually tried include training epoch ($10$, $20$), learning rate ($1e-3$,$1e-4$ for the classifier head, $1e-4$, $5e-5$, $1e-5$ for BERT, note that the learning rate of the classifier head is always larger than BERT). The ablation study and the analysis on radius \(r\) can be found in \Figref{fig:exp_radius} and \Appenref{app:sec:radius}.
% Considering that the improvement by adding gradient ascend is not as impressive as projection (Check ~\Tableref{table:discussion_negative}).
The hyper-parameters in gradient ascend are not sensitive to the final experiments.

\subsection{More results}
\paragraph{F1-known and F1-open}

\input{tabels/app_open_known}
The definition of F1-known and F1-open can be found in \Secref{sec:exp:para:eval}. \Tableref{app:table:main} shows the comparisons between the baselines and ours.

\paragraph{Reproducibility Checklist: Differences on Datasets}

During this process, we found that dataset CLINC is the robustest to change of hyper-parameters while dataset Stackoverflow is the weakest. A similar observation also shows in \citet{zeng2021modeling}. It works the best on CLINC and worst on Stackoverflow.

We hypothesize that the difference comes from the quality of the raw data. As shown in Table~\Tableref{app:table:demo}, the category of the input in Stackoverflow is usually included in the original sentence and we name them ``easy''. Rare sentences do not follow to this rule and we call them ``hard''. This leads to an observation in empirical experiments that the number of training epochs should be controlled in a limited range; otherwise many open category samples would be wrongly categorized to the known category. 
% Limited by the page, we did not 

This is in consistent with the finding in noisy labeling classification. The neural network will first fit the clean label before overfitting the noisy labeled samples. Under the current setting, the ``hard'' corresponds to the noisy sample. The \ovr will first fit the easier one, followed by the harder one. When the harder one is classified correctly, many open categories could also be classified into this known category. 

ADB~\citep{zhang2021deep} avoids this problem by working directly on the pre-trained features. It can statistically filter out the influence of noisy samples. Though ADB does not require extra hyper-parameter tuning, we found that the position of features extracted from the model has an impact on the final performances.  
% As a result, ADB is more robust to hyper-parameters on Stackoverflow. 

In summary, the differences between datasets are an intriguing topic that merits further investigation in the future.

% Because of this, the training 
% \paragraph{Reproducibility Checklist: performance on validation set}

% \paragraph{Vanilla \ovr framework}
% what? 
% \paragraph{Special case}

\paragraph{Reproducibility Checklist: Standard Deviation}
\input{tabels/main_results_std}
As shown in \Tableref{app:table:main_std}, larger known category ratios are more likely to be associated with lower variance; this is to be expected because more samples make the training more stable.

\subsection{Ablation Study: Synthesized Negative Samples on other Structures}
Negative sample generation for MSP and ADB follow the same process as ours, except that the gradient ascend is removed. The complete version is left for future work.

\label{app:sec:ans}
\paragraph{MSP with negative sampling}
The original MSP is a C-way classifier \(f^{cls}\) trained with cross-entropy loss during the training. In inference, the confidence score \(p(\xv_i) = \text{Softmax}(f^{cls}(\xv_i))\) is first calculated. If the category with maximum probability \(p_m\) is lower than \(0.5\), the corresponding input is recognized as ``open'' category. Otherwise, this sample belongs to the category \(m\), \ie, 
\begin{align*}
\hat{y}=
\begin{cases}
	\text{open,} & \text{if } \max(p(\xv_i)) < 0.5 \\
	 \arg \max p(\xv_i) , & \text{ otherwise}.
\end{cases}
% 	\end{equation}
\end{align*}
% \paragraph{MSP with negative}

In MSP with negative settings, an extra category \(l_0\) is added for synthesized negatives. The inference now becomes
\begin{align*}
\hat{y}=
\begin{cases}
	\text{open,} & \text{if } \max(p(\xv_i)) < 0.5 \\
	 \text{open,} & \text{if } \argmax(p(\xv_i)) == l_0 \\
	 \arg \max p(\xv_i) , & \text{ otherwise} \\
\end{cases}
% 	\end{equation}
\end{align*}
% where \(l_0\) represents the 
Note that our MSP with synthesized negatives differs from ~\citep{zhan2021out} in two aspects, (\(i\)), different ways to choose the negative samples. \(ii\)), their work added synthesized negative samples to the validation set, while ours uses the origin validation set.

\paragraph{ADB with negative sampling}
ADB training has two steps. The first step is to learn a good feature extractor using C-way classifier. The second step is to learn the boundary of each category in the pre-trained feature space. 

Our modification is the first step. We replace the origin classifier with a \(C+1\)-way classifier. The extra head is designed for the synthesized negative samples. The inference step is kept the same as the original method.

% \subsection{Ablation Study: Synthesized Negative Samples Generation}

% \subsection{Code}
% Please find it in supplementary.

%% file: tabels/app_dataset_domo.tex
\begin{table*}[]
\begin{center}
% \begin{sc}
   \scalebox{0.73}{
\begin{tabular}{@{}c|c|l@{}}
\toprule
Dataset                        & Category                                                                                   & \multicolumn{1}{c}{Examples}                                                                                                                                                                                                                      \\ \midrule
\multirow{3}{*}{Banking}       & exchange\_via\_app                                                                         & \begin{tabular}[c]{@{}l@{}}What currences are available for exchange?\\ Does your app allow currency exchange from USD to GBP?\\ I want to exchange USD and GBP with the app\end{tabular}                                                         \\ \cmidrule(l){2-3} 
                               & \begin{tabular}[c]{@{}c@{}}wrong\_exchange\_\\ rate\_for\_cash\\ \_withdrawal\end{tabular} & \begin{tabular}[c]{@{}l@{}}I was given the wrong exchange rate when getting cash\\ I think I was charged a different exchange rate than what was posted at the time.\\ I need an accurate exchange rate, when I make my withdrawals.\end{tabular} \\ \cmidrule(l){2-3} 
                               & \begin{tabular}[c]{@{}c@{}}card\_payment\_\\ wrong\_exchange\_rate\end{tabular}            & \begin{tabular}[c]{@{}l@{}}Why didn't I receive the correct exchange rate for an item that I purchased?\\ The fee charged when I changed rubles into British pounds was too much.\\ I am being charged the wrong amount on my card.\end{tabular}  \\ \midrule
\multirow{3}{*}{CLINC}        & flight\_status                                                                             & \begin{tabular}[c]{@{}l@{}}what time is this flight supposed to land\\ what time will i be able to board the plane\\ so when is my flight landing\end{tabular}                                                                                    \\ \cmidrule(l){2-3} 
                               & time                                                                                       & \begin{tabular}[c]{@{}l@{}}what time is it in adelaide, australia right now\\ please tell me the time\\ how late is it now in ourense\end{tabular}                                                                                                \\ \cmidrule(l){2-3} 
                               & how\_busy                                                                                  & \begin{tabular}[c]{@{}l@{}}how long will i wait for a table at red lobster\\ can i expect chili's to be busy at 4:30\\ so how busy is the outback steakhouse at 5 pm\end{tabular}                                                                 \\ \midrule
\multirow{3}{*}{Stackoverflow} & wordpress                                                                                  & \begin{tabular}[c]{@{}l@{}}How Display Recent Posts in all 3 languages at once in Wordpress\\ How secure is Wordpress?\\ Need MySQL Queries to delete WordPress Posts and Post Meta more than X Days Old\end{tabular}                             \\ \cmidrule(l){2-3} 
                               & apache                                                                                     & \begin{tabular}[c]{@{}l@{}}Apache / PHP Disable Cookies for Subdomain ?\\ Increase PHP Memory limit (Apache, Drupal6)\\ is setting the uploads folder 777 permision secure\end{tabular}                                                           \\ \cmidrule(l){2-3} 
                               & excel                                                                                      & \begin{tabular}[c]{@{}l@{}}Condition to check whether cell is readonly in EXCEL using C\#\\ EXCEL XOR multiple bits\\ How I can export a datatable to excel 2007 and pdf from asp.net?\end{tabular}                                               \\ \bottomrule
\end{tabular}
}
% \end{sc}
\end{center}
\caption{Extracted samples from three main datasets.}
\label{app:table:demo}
\end{table*}

%% file: tabels/dataset.tex
% \usepackage{booktabs}
\begin{table}[t]{}
\begin{center}
\begin{sc}
   \scalebox{0.7}{
% \begin{minipage}[t]{0.4\textwidth}
\begin{tabular}{@{}cccc@{}}
\toprule
Dataset       & \begin{tabular}[c]{@{}c@{}}Avg. Samples \\ per category\end{tabular} & Avg. Length & Classes \\ \midrule
Banking       & 117                                                                  & 11.91       & 77      \\
CLINIC     & 100                                                                  & 8.31        & 150     \\
StackOverflow & 600                                                                  & 9.18        & 20      \\
% 20NewsGroup   & 660                                                                  & 228.82      & 20      \\
\bottomrule
\end{tabular}}
\end{sc}
\end{center}
% \end{minipage}
\caption{Statistics of benchmark datasets}
\label{tab:dataset_stats}
\end{table}

%% file: tabels/app_open_known.tex
\begin{table*}[ht!]
\begin{center}
\begin{sc}
   \scalebox{0.9}{
\begin{tabular}{cccccccc}
\hline
\multicolumn{1}{l}{} & \multicolumn{1}{l}{} & \multicolumn{2}{c}{BANKING}     & \multicolumn{2}{c}{CLINIC}      & \multicolumn{2}{c}{StackOverflow} \\ \hline
\%                  & Methods              & Open           & Known          & Open           & Known          & Open            & Known           \\ \hline
\multirow{7}{*}{25}  & MSP                  & 41.43          & 50.55          & 50.88          & 47.53          & 13.03           & 42.82           \\
                     & DOC                  & 61.42          & 57.85          & 81.98          & 65.96          & 41.25           & 49.02           \\
                     & OpenMax              & 51.32          & 54.28          & 75.76          & 61.62          & 36.41           & 47.89           \\
                     & DeepUnk              & 70.44          & 60.88          & 87.33          & 70.73          & 49.29           & 52.60           \\
                     & ADB                  & 84.56          & 70.94          & 91.84          & 76.80          & 90.88           & 78.82           \\
                     & SelfSup*             & 80.12          & 69.39          & 92.35          & 80.43          & 74.86           & 63.80           \\
                     & Ours                 & \textbf{86.39} & \textbf{72.29} & \textbf{95.33} & \textbf{84.53} & \textbf{93.96}  & \textbf{82.63}  \\ \hline
\multirow{7}{*}{50}  & MSP                  & 41.19          & 71.97          & 57.62          & 70.58          & 23.99           & 66.91           \\
                     & DOC                  & 55.14          & 73.59          & 79.00          & 78.25          & 25.44           & 66.58           \\
                     & OpenMax              & 54.33          & 74.76          & 81.89          & 80.54          & 45.00           & 70.49           \\
                     & DeepUnk              & 69.53          & 77.74          & 85.85          & 82.11          & 43.01           & 70.51           \\
                     & ADB                  & 78.44          & 80.96          & 88.65          & 85.00          & \textbf{87.34}  & {85.68}  \\
                     & SelfSup*             & 67.26          & 79.52          & 90.30          & 86.54          & 71.88           & 79.22           \\
                     & Ours                 & \textbf{81.06} & \textbf{83.52} & \textbf{92.16} & \textbf{87.95} & 86.83           & \textbf{85.81}           \\ \hline
\multirow{7}{*}{75}  & MSP                  & 39.23          & 84.36          & 59.08          & 82.59          & 33.96           & 80.88           \\
                     & DOC                  & 50.60          & 83.91          & 72.87          & 83.69          & 16.76           & 78.95           \\
                     & OpenMax              & 50.85          & 84.64          & 76.35          & 73.13          & 44.87           & 82.11           \\
                     & DeepUnk              & 58.54          & 84.75          & 81.15          & 86.27          & 37.59           & 81.00           \\
                     & ADB                  & 66.47          & 86.29          & 83.92          & 88.58          & 73.86  & 86.80           \\
                     & SelfSup*             & 60.71          & 87.47          & 86.28          & \textbf{89.46} & 65.44           & 87.22           \\
                     & Ours                 & \textbf{70.54} & \textbf{88.13}          & \textbf{87.20} & 89.18          & \textbf{74.82}           & \textbf{88.34}  \\ \hline
\end{tabular}
}
\end{sc}
\end{center}
\caption{Macro-F1 score on known category and open category of open world classification on three datasets with different known class proportions. 
% The scores of baseline methods are from the~\citep{zhang2021deep} and ~\citep{zhan2021out}.
* means extra datasets are used during the training. This table complements to Table~\ref{table:main}}
\label{app:table:main}
\end{table*}

%% file: tabels/main_results_std.tex
\begin{table*}[ht!]
\begin{center}
\begin{sc}
   \scalebox{0.9}{
\begin{tabular}{@{}cccccc@{}}
\toprule
\multicolumn{1}{l}{}           & \% & Accuracy & F1   & F1-open & F1-known \\ \midrule
\multirow{3}{*}{Banking}       & 25 & 1.83     & 1.76 & 1.47    & 1.78     \\
                               & 50 & 1.05     & 1.00 & 1.14    & 1.00     \\
                               & 75 & 1.05     & 0.69 & 2.45    & 0.67     \\ \midrule
\multirow{3}{*}{CLINC}         & 25 & 1.85     & 2.60 & 1.25    & 2.64     \\
                               & 50 & 0.86     & 0.74 & 0.75    & 0.74     \\
                               & 75 & 0.68     & 0.43 & 0.91    & 0.42     \\ \midrule
\multirow{3}{*}{StackOverflow} & 25 & 1.48     & 2.01 & 1.02    & 2.22     \\
                               & 50 & 2.85     & 2.31 & 3.13    & 2.25     \\
                               & 75 & 1.21     & 0.54 & 3.10    & 0.43     \\ \bottomrule
\end{tabular}
}
\end{sc}
\end{center}
\caption{Standard deviation of results collected from different random seeds. This table complements to Table~\ref{table:main}}
\label{app:table:main_std}
\end{table*}

%% file: emnlp2022.bbl
\begin{thebibliography}{37}
\expandafter\ifx\csname natexlab\endcsname\relax\def\natexlab#1{#1}\fi

\bibitem[{Bendale and Boult(2015)}]{bendale2015towards}
Abhijit Bendale and Terrance Boult. 2015.
\newblock Towards open world recognition.
\newblock In \emph{Proceedings of the IEEE conference on computer vision and
  pattern recognition}, pages 1893--1902.

\bibitem[{Bendale and Boult(2016)}]{bendale2016towards}
Abhijit Bendale and Terrance~E Boult. 2016.
\newblock Towards open set deep networks.
\newblock In \emph{Proceedings of the IEEE conference on computer vision and
  pattern recognition}, pages 1563--1572.

\bibitem[{Bengio et~al.(2013)Bengio, Mesnil, Dauphin, and
  Rifai}]{bengio2013better}
Yoshua Bengio, Gr{\'e}goire Mesnil, Yann Dauphin, and Salah Rifai. 2013.
\newblock Better mixing via deep representations.
\newblock In \emph{International conference on machine learning}, pages
  552--560. PMLR.

\bibitem[{Boyd et~al.(2004)Boyd, Boyd, and Vandenberghe}]{boyd2004convex}
Stephen Boyd, Stephen~P Boyd, and Lieven Vandenberghe. 2004.
\newblock \emph{Convex optimization}.
\newblock Cambridge university press.

\bibitem[{Breunig et~al.(2000)Breunig, Kriegel, Ng, and
  Sander}]{breunig2000lof}
Markus~M Breunig, Hans-Peter Kriegel, Raymond~T Ng, and J{\"o}rg Sander. 2000.
\newblock Lof: identifying density-based local outliers.
\newblock In \emph{Proceedings of the 2000 ACM SIGMOD international conference
  on Management of data}, pages 93--104.

\bibitem[{Casanueva et~al.(2020)Casanueva, Tem{\v{c}}inas, Gerz, Henderson, and
  Vuli{\'c}}]{casanueva2020efficient}
Inigo Casanueva, Tadas Tem{\v{c}}inas, Daniela Gerz, Matthew Henderson, and
  Ivan Vuli{\'c}. 2020.
\newblock Efficient intent detection with dual sentence encoders.
\newblock \emph{arXiv preprint arXiv:2003.04807}.

\bibitem[{Chen et~al.(2020)Chen, Kornblith, Norouzi, and
  Hinton}]{chen2020simple}
Ting Chen, Simon Kornblith, Mohammad Norouzi, and Geoffrey Hinton. 2020.
\newblock A simple framework for contrastive learning of visual
  representations.
\newblock In \emph{International conference on machine learning}, pages
  1597--1607. PMLR.

\bibitem[{Devlin et~al.(2018)Devlin, Chang, Lee, and
  Toutanova}]{devlin2018bert}
Jacob Devlin, Ming-Wei Chang, Kenton Lee, and Kristina Toutanova. 2018.
\newblock Bert: Pre-training of deep bidirectional transformers for language
  understanding.
\newblock \emph{arXiv preprint arXiv:1810.04805}.

\bibitem[{Goyal et~al.(2020)Goyal, Raghunathan, Jain, Simhadri, and
  Jain}]{goyal2020drocc}
Sachin Goyal, Aditi Raghunathan, Moksh Jain, Harsha~Vardhan Simhadri, and
  Prateek Jain. 2020.
\newblock Drocc: Deep robust one-class classification.
\newblock In \emph{International Conference on Machine Learning}, pages
  3711--3721. PMLR.

\bibitem[{Guo et~al.(2017)Guo, Pleiss, Sun, and
  Weinberger}]{guo2017calibration}
Chuan Guo, Geoff Pleiss, Yu~Sun, and Kilian~Q Weinberger. 2017.
\newblock On calibration of modern neural networks.
\newblock In \emph{International Conference on Machine Learning}, pages
  1321--1330. PMLR.

\bibitem[{He et~al.(2020)He, Fan, Wu, Xie, and Girshick}]{he2020momentum}
Kaiming He, Haoqi Fan, Yuxin Wu, Saining Xie, and Ross Girshick. 2020.
\newblock Momentum contrast for unsupervised visual representation learning.
\newblock In \emph{Proceedings of the IEEE/CVF conference on computer vision
  and pattern recognition}, pages 9729--9738.

\bibitem[{Hendrycks and Gimpel(2016)}]{hendrycks2016baseline}
Dan Hendrycks and Kevin Gimpel. 2016.
\newblock A baseline for detecting misclassified and out-of-distribution
  examples in neural networks.
\newblock \emph{arXiv preprint arXiv:1610.02136}.

\bibitem[{Larson et~al.(2019)Larson, Mahendran, Peper, Clarke, Lee, Hill,
  Kummerfeld, Leach, Laurenzano, Tang et~al.}]{larson2019evaluation}
Stefan Larson, Anish Mahendran, Joseph~J Peper, Christopher Clarke, Andrew Lee,
  Parker Hill, Jonathan~K Kummerfeld, Kevin Leach, Michael~A Laurenzano,
  Lingjia Tang, et~al. 2019.
\newblock An evaluation dataset for intent classification and out-of-scope
  prediction.
\newblock \emph{arXiv preprint arXiv:1909.02027}.

\bibitem[{Lewis et~al.(2019)Lewis, Liu, Goyal, Ghazvininejad, Mohamed, Levy,
  Stoyanov, and Zettlemoyer}]{lewis2019bart}
Mike Lewis, Yinhan Liu, Naman Goyal, Marjan Ghazvininejad, Abdelrahman Mohamed,
  Omer Levy, Ves Stoyanov, and Luke Zettlemoyer. 2019.
\newblock Bart: Denoising sequence-to-sequence pre-training for natural
  language generation, translation, and comprehension.
\newblock \emph{arXiv preprint arXiv:1910.13461}.

\bibitem[{Liang et~al.(2018)Liang, Li, and Srikant}]{liang2017enhancing}
Shiyu Liang, Yixuan Li, and Rayadurgam Srikant. 2018.
\newblock Enhancing the reliability of out-of-distribution image detection in
  neural networks.
\newblock \emph{ICLR}.

\bibitem[{Lin and Xu(2019{\natexlab{a}})}]{lin2019deep}
Ting-En Lin and Hua Xu. 2019{\natexlab{a}}.
\newblock Deep unknown intent detection with margin loss.
\newblock \emph{arXiv preprint arXiv:1906.00434}.

\bibitem[{Lin and Xu(2019{\natexlab{b}})}]{lin2019post}
Ting-En Lin and Hua Xu. 2019{\natexlab{b}}.
\newblock A post-processing method for detecting unknown intent of dialogue
  system via pre-trained deep neural network classifier.
\newblock \emph{Knowledge-Based Systems}, 186:104979.

\bibitem[{Madry et~al.(2018)Madry, Makelov, Schmidt, Tsipras, and
  Vladu}]{Madry2018TowardsDL}
Aleksander Madry, Aleksandar Makelov, Ludwig Schmidt, Dimitris Tsipras, and
  Adrian Vladu. 2018.
\newblock Towards deep learning models resistant to adversarial attacks.
\newblock \emph{ArXiv}, abs/1706.06083.

\bibitem[{Miyato et~al.(2016)Miyato, Dai, and
  Goodfellow}]{miyato2016adversarial}
Takeru Miyato, Andrew~M Dai, and Ian Goodfellow. 2016.
\newblock Adversarial training methods for semi-supervised text classification.
\newblock \emph{arXiv preprint arXiv:1605.07725}.

\bibitem[{Pless and Souvenir(2009)}]{pless2009survey}
Robert Pless and Richard Souvenir. 2009.
\newblock A survey of manifold learning for images.
\newblock \emph{IPSJ Transactions on Computer Vision and Applications},
  1:83--94.

\bibitem[{Rifkin and Klautau(2004)}]{rifkin2004defense}
Ryan Rifkin and Aldebaro Klautau. 2004.
\newblock In defense of one-vs-all classification.
\newblock \emph{The Journal of Machine Learning Research}, 5:101--141.

\bibitem[{Ruff et~al.(2018)Ruff, Vandermeulen, Goernitz, Deecke, Siddiqui,
  Binder, M{\"u}ller, and Kloft}]{ruff2018deep}
Lukas Ruff, Robert Vandermeulen, Nico Goernitz, Lucas Deecke, Shoaib~Ahmed
  Siddiqui, Alexander Binder, Emmanuel M{\"u}ller, and Marius Kloft. 2018.
\newblock Deep one-class classification.
\newblock In \emph{International conference on machine learning}, pages
  4393--4402. PMLR.

\bibitem[{Scheirer et~al.(2012)Scheirer, de~Rezende~Rocha, Sapkota, and
  Boult}]{scheirer2012toward}
Walter~J Scheirer, Anderson de~Rezende~Rocha, Archana Sapkota, and Terrance~E
  Boult. 2012.
\newblock Toward open set recognition.
\newblock \emph{IEEE transactions on pattern analysis and machine
  intelligence}, 35(7):1757--1772.

\bibitem[{Scheirer et~al.(2014)Scheirer, Jain, and
  Boult}]{scheirer2014probability}
Walter~J Scheirer, Lalit~P Jain, and Terrance~E Boult. 2014.
\newblock Probability models for open set recognition.
\newblock \emph{IEEE transactions on pattern analysis and machine
  intelligence}, 36(11):2317--2324.

\bibitem[{Sch{\"o}lkopf et~al.(2001)Sch{\"o}lkopf, Platt, Shawe-Taylor, Smola,
  and Williamson}]{scholkopf2001estimating}
Bernhard Sch{\"o}lkopf, John~C Platt, John Shawe-Taylor, Alex~J Smola, and
  Robert~C Williamson. 2001.
\newblock Estimating the support of a high-dimensional distribution.
\newblock \emph{Neural computation}, 13(7):1443--1471.

\bibitem[{Shu et~al.(2021)Shu, Benajiba, Mansour, and Zhang}]{shu2021odist}
Lei Shu, Yassine Benajiba, Saab Mansour, and Yi~Zhang. 2021.
\newblock Odist: Open world classification via distributionally shifted
  instances.
\newblock In \emph{Findings of the Association for Computational Linguistics:
  EMNLP 2021}, pages 3751--3756.

\bibitem[{Shu et~al.(2017)Shu, Xu, and Liu}]{shu2017doc}
Lei Shu, Hu~Xu, and Bing Liu. 2017.
\newblock Doc: Deep open classification of text documents.
\newblock \emph{arXiv preprint arXiv:1709.08716}.

\bibitem[{Wang et~al.(2018)Wang, Wang, Zhou, Ji, Gong, Zhou, Li, and
  Liu}]{wang2018cosface}
Hao Wang, Yitong Wang, Zheng Zhou, Xing Ji, Dihong Gong, Jingchao Zhou, Zhifeng
  Li, and Wei Liu. 2018.
\newblock Cosface: Large margin cosine loss for deep face recognition.
\newblock In \emph{Proceedings of the IEEE conference on computer vision and
  pattern recognition}, pages 5265--5274.

\bibitem[{Wang et~al.(2021)Wang, Li, Che, Zhou, Liu, and Li}]{wang2021energy}
Yezhen Wang, Bo~Li, Tong Che, Kaiyang Zhou, Ziwei Liu, and Dongsheng Li. 2021.
\newblock Energy-based open-world uncertainty modeling for confidence
  calibration.
\newblock In \emph{Proceedings of the IEEE/CVF International Conference on
  Computer Vision}, pages 9302--9311.

\bibitem[{Xu et~al.(2015)Xu, Wang, Tian, Xu, Zhao, Wang, and Hao}]{xu2015short}
Jiaming Xu, Peng Wang, Guanhua Tian, Bo~Xu, Jun Zhao, Fangyuan Wang, and
  Hongwei Hao. 2015.
\newblock Short text clustering via convolutional neural networks.
\newblock In \emph{Proceedings of the 1st Workshop on Vector Space Modeling for
  Natural Language Processing}, pages 62--69.

\bibitem[{Yan et~al.(2020)Yan, Fan, Li, Liu, Zhang, Wu, and
  Lam}]{yan2020unknown}
Guangfeng Yan, Lu~Fan, Qimai Li, Han Liu, Xiaotong Zhang, Xiao-Ming Wu, and
  Albert~YS Lam. 2020.
\newblock Unknown intent detection using gaussian mixture model with an
  application to zero-shot intent classification.
\newblock In \emph{Proceedings of the 58th Annual Meeting of the Association
  for Computational Linguistics}, pages 1050--1060.

\bibitem[{Zeng et~al.(2021)Zeng, He, Yan, Liu, Wu, Xu, Jiang, and
  Xu}]{zeng2021modeling}
Zhiyuan Zeng, Keqing He, Yuanmeng Yan, Zijun Liu, Yanan Wu, Hong Xu, Huixing
  Jiang, and Weiran Xu. 2021.
\newblock Modeling discriminative representations for out-of-domain detection
  with supervised contrastive learning.
\newblock \emph{arXiv preprint arXiv:2105.14289}.

\bibitem[{Zhan et~al.(2021)Zhan, Liang, Liu, Fan, Wu, and Lam}]{zhan2021out}
Li-Ming Zhan, Haowen Liang, Bo~Liu, Lu~Fan, Xiao-Ming Wu, and Albert Lam. 2021.
\newblock Out-of-scope intent detection with self-supervision and
  discriminative training.
\newblock \emph{arXiv preprint arXiv:2106.08616}.

\bibitem[{Zhang et~al.(2021{\natexlab{a}})Zhang, Li, Xu, Zhang, Zhao, and
  Gao}]{zhang-etal-2021-textoir}
Hanlei Zhang, Xiaoteng Li, Hua Xu, Panpan Zhang, Kang Zhao, and Kai Gao.
  2021{\natexlab{a}}.
\newblock {TEXTOIR}: An integrated and visualized platform for text open intent
  recognition.
\newblock In \emph{Proceedings of the 59th Annual Meeting of the Association
  for Computational Linguistics and the 11th International Joint Conference on
  Natural Language Processing: System Demonstrations}, pages 167--174.

\bibitem[{Zhang et~al.(2021{\natexlab{b}})Zhang, Xu, and Lin}]{zhang2021deep}
Hanlei Zhang, Hua Xu, and Ting-En Lin. 2021{\natexlab{b}}.
\newblock Deep open intent classification with adaptive decision boundary.
\newblock In \emph{Proceedings of the AAAI Conference on Artificial
  Intelligence}, volume~35, pages 14374--14382.

\bibitem[{Zhou et~al.(2022)Zhou, Liu, and Qiu}]{zhou2022knn}
Yunhua Zhou, Peiju Liu, and Xipeng Qiu. 2022.
\newblock Knn-contrastive learning for out-of-domain intent classification.
\newblock In \emph{Proceedings of the 60th Annual Meeting of the Association
  for Computational Linguistics (Volume 1: Long Papers)}, pages 5129--5141.

\bibitem[{Zhu et~al.(2019)Zhu, Cheng, Gan, Sun, Goldstein, and
  Liu}]{zhu2019freelb}
Chen Zhu, Yu~Cheng, Zhe Gan, Siqi Sun, Tom Goldstein, and Jingjing Liu. 2019.
\newblock Freelb: Enhanced adversarial training for natural language
  understanding.
\newblock \emph{ICLR}.

\end{thebibliography}
